\documentclass[final,12pt,3p]{elsarticle}

\usepackage{amsmath,amssymb}

\DeclareMathOperator{\EX}{\mathbb{E}}

\usepackage{booktabs}
\usepackage{longtable}
\usepackage{lscape}
\usepackage{hyperref}

\usepackage{lineno}
\modulolinenumbers[5]
\usepackage{amssymb}
\usepackage{amsmath,eurosym}
\usepackage{gensymb}
\usepackage{calrsfs}
\usepackage{bbding} 
\usepackage{rotating}
\usepackage{tabulary}   
\usepackage{tabularx}
\usepackage{adjustbox}
\usepackage{graphicx}
\usepackage{multirow}
\usepackage{float}
\usepackage{multicol}
\usepackage{nomencl}
\usepackage[table]{xcolor}
\usepackage{pifont}                     
\newcommand{\xmark}{\ding{55}}
\usepackage{siunitx}
\usepackage[ruled,vlined]{algorithm2e}

\usepackage{etoolbox}
\usepackage{nomencl}                    
\newcommand\Nomenclature[3][X]{\nomenclature[#1#3]{#2}{#3}}

\biboptions{comma,square,sort&compress}

\usepackage{caption}
\usepackage{subcaption}

\usepackage{todonotes}
\newcommand{\tinytodo}[2][]{\todo[caption={#2}, size=\small, #1]{\renewcommand{\baselinestretch}{0.5}\selectfont#2\par}}
\newcolumntype{M}[1]{>{\centering\arraybackslash}m{#1}}
\newcommand\hlr[1]{%
  \bgroup
  \hskip0pt\color{red!80!black}%
  #1%
  \egroup}
 \renewcommand{\tinytodo}[2][]{}
 \renewcommand\hlr[1]{#1}

\newcolumntype{M}[1]{>{\centering\arraybackslash}m{#1}}
\newcommand\hlrrevnew[1]{%
  \bgroup
  \hskip0pt\color{red!80!black}%
  #1%
  \egroup}
 
\usepackage{marginnote}

\journal{Energy and AI}

\begin{document}

\begin{frontmatter}


\title{Development of a Soft Actor Critic Deep Reinforcement Learning Approach for Harnessing Energy Flexibility in a Large Office Building}

\author[label1,label2]{Anjukan Kathirgamanathan\corref{cor1}}
\address[label1]{School of Mechanical and Materials Engineering, University College Dublin}
\address[label2]{UCD Energy Institute, O'Brien Centre for Science, University College Dublin\fnref{label4}}

\cortext[cor1]{Corresponding author}

\ead{anjukan.kathirgamanathan@ucdconnect.ie}

\author[label2,label5]{Eleni Mangina}
\address[label5]{School of Computer Science, University College Dublin}

\author[label1,label2]{Donal P. Finn}


\begin{abstract}

This research is concerned with the novel application and investigation of `Soft Actor Critic' (SAC) based Deep Reinforcement Learning (DRL) to control the cooling setpoint (and hence cooling loads) of a large commercial building to harness energy flexibility. The research is motivated by the challenge associated with the development and application of conventional model-based control approaches at scale to the wider building stock. SAC is a model-free DRL technique that is able to handle continuous action spaces and which has seen limited application to real-life or high-fidelity simulation implementations in the context of automated and intelligent control of building energy systems. Such control techniques are seen as one possible solution to supporting the operation of a smart, sustainable and future electrical grid. This research tests the suitability of the SAC DRL technique through training and deployment of the agent on an EnergyPlus based environment of the office building. The SAC DRL was found to learn an optimal control policy that was able to minimise energy costs by 9.7\% compared to the default rule-based control (RBC) scheme and was able to improve or maintain thermal comfort limits over a test period of one week. The algorithm was shown to be robust to the different hyperparameters and this optimal control policy was learnt through the use of a minimal state space consisting of readily available variables. The robustness of the algorithm was tested through investigation of the speed of learning and ability to deploy to different seasons and climates. It was found that the SAC DRL requires minimal training sample points and outperforms the RBC after three months of operation and also without disruption to thermal comfort during this period. The agent is transferable to other climates and seasons although further retraining or hyperparameter tuning is recommended.
\newline
\newline
\textit{Highlights:}
\begin{itemize}
    \item A novel application of Soft Actor Critic Deep Reinforcement Learning (SAC DRL).
    \item Controller harnesses energy flexibility from building passive thermal mass.
    \item A novel investigation into robustness of hyperparameters and state space design.
    \item SAC DRL able to achieve cost savings of 9.7\% compared to the baseline RBC.
    \item Minimal training is required without the need for disruptive excitation.
\end{itemize}

\end{abstract}


\end{frontmatter}


\section{Introduction}
\label{sec1}
\subsection{The Importance of Building Energy Flexibility in the Smart Grid}
\label{sec1.1}
Renewable energy sources such as wind and solar are intrinsically variable by nature and this has the potential to create a stability challenge for the grid with the fluctuating supply needing to be balanced with demand \cite{Lund2015}. In their review of the future of the low-carbon electricity grid, \citet{Greenblatt2017} critique such conventional renewable energy technologies. They suggest that these technologies alone do not represent an ideal solution and that grid integration is required in conjunction to deliver a reliable and robust electricity system. \citet{Holttinen2009} focus on the impacts of large amounts of wind power on the design and operation of power systems. \citet{Villar2018} summarise some of the challenges faced by this new power system and the need for new flexibility products and markets. The flexibility to manage any mismatch can come from either the supply side (through the use of dedicated conventional power plants or storage) or from the demand side \cite{Lund2015, Cochran2014}. Demand Side Management (DSM) is one such grid integration strategy and can be broadly categorised as actions that influence the quantity, patterns of use or the primary source of energy consumed by end users \cite{Hull2012}. Demand Response (DR) is one promising facet of DSM where consumers curtail or shift their electricity usage in response to financial or other incentives. Within DR, there are different strategies based on the response times, services offered and business models \cite{Klein2017}.

With buildings representing about 40\% of the total primary energy consumption in Europe \cite{Economidou2011}, they are very relevant to participation in DR and the provision of energy flexibility. Further, the thermal mass of buildings allows them to be used as a thermal energy storage system making them potentially very useful in DSM \cite{Reynders2013}. Commercial buildings are of particular interest given their greater thermal mass and common usage of space conditioning \cite{Aduda2017}. This is often through the use of heating, ventilation and air conditioning (HVAC) systems and this HVAC load is one such load that can be shifted using the thermal mass of the building. These HVAC systems are often integrated with Building Automation Systems (BAS) or Building Energy Management Systems (BEMS) which can be used to automate DR measures. These systems are also capable of receiving signals directly from the electricity grid or aggregator \cite{Hao2012}. Buildings may often also possess active thermal storage, active electric storage (batteries), indirect electric storage (Electric Vehicles) and on-site generation as further sources of energy flexibility.

DR programs can be categorised as being either price or incentive based \cite{Vardakas2015}. Incentive-based programmes pay customers (end-users) to shift their electricity consumption at times requested by grid operators. Grid operators can be either distribution system operators, who are generally responsible for the operation of the low-voltage distribution system and delivery of power to the end consumers; or transmission system operators, who are responsible for the operation of the high-voltage transmission system and ensuring its stability. Generally, a building is required to be capable of meeting a minimum required reduction in power consumption. While individual buildings may not be capable of meeting this reduction, aggregators are a market actor who contract with these buildings and combine the available power reduction and offer this to the grid operator, receiving a percentage of the value to the grid operator created by applying the DSM measure \cite{Jensen2017}. Given that energy flexibility is a resource that is aggregated from many buildings, which are all unique in the way that they are constructed, designed and operated, a scalable and transferable method of assessing and harnessing this energy flexibility is required \cite{Kathirgamanathan2020}. Artificial intelligence and machine learning techniques have gained prominence in a wide variety of applications and domains in recent times. Reinforcement learning has emerged as a promising automated high-level control technique for buildings \cite{Vazquez-Canteli2019, Wang2020, Azuatalam2020, Mason2019}, that is a potentially model-free control approach avoiding the need to create non-generalisable models of the building.

\subsection{Reinforcement Learning Applied to Building Energy Management}
\label{sec1.2}

Reinforcement learning (RL) is an area of machine learning, where an agent learns to take the optimal set of actions through interaction in a dynamic environment (such as a building subject to changing weather conditions, varying grid requirements and occupants with thermal comfort requirements), with the goal of maximising a certain reward quantity \citep{Mason2019}. For a comprehensive introduction to the field of RL, the reader is referred to standard textbooks \citep{Sutton2014}. A brief introduction to RL is provided here with a formal definition and a description of the different types of RL algorithms. This section is closed with a critical analysis of the approach and its suitability for DR applications.

\begin{figure}[!htbp]
\centering
\includegraphics[clip, trim=7cm 7.5cm 8cm 6.5cm]{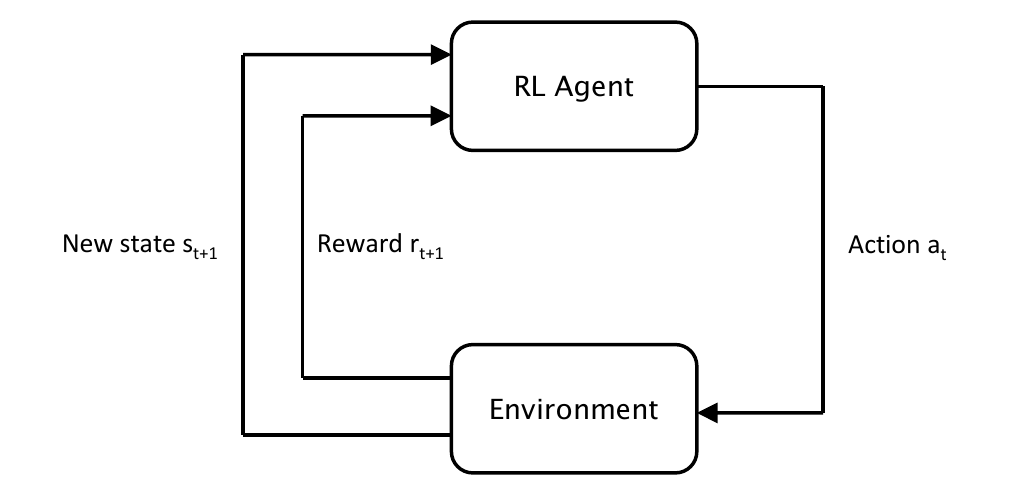}
\caption{Agent-environment interaction in reinforcement learning}
\label{fig:2_RL}
\end{figure}

Given a state, for every action that the agent takes (either discrete or continuous action space), this leads to a new state in the environment and based on this, the agent is either rewarded or penalised for taking that particular action (see Figure \ref{fig:2_RL}). The reward is a feedback mechanism to the agent to indicate how well it is performing at each time step. The agent has the objective of maximising the cumulative reward until a terminal condition is reached for the episode. In this way, the RL problem can be formalised as a Markov Decision Process (MDP) containing four elements:
\begin{enumerate}
    \item $S$, a set of states (e.g., zone indoor temperature, outdoor relative humidity)
    \item $A$, a set of actions (e.g., zone set point temperatures)
    \item $r: S \times A$, a reward function describing the reward as a result of taking a specific action
    \item $P: S \times A \times S' \in [0,1]$, transition probabilities between the states
\end{enumerate}
For the state $S_t$ to satisfy the Markov condition, the future state must only be dependent on the current state and current actions, i.e., the future state is independent of the past state, given the present state \citep{Sutton2014}. 

The major components of an RL agent are policy, return and value functions \citep{Azuatalam2020}. The policy $\pi$ maps states to actions as $\pi: S \rightarrow A$. A value function ($V^{\pi}(S)$) can be defined for a given state $s$ as follows:

\begin{equation}
\label{eq:2_bellman}
  V^{\pi}(S) = \sum\limits_{a}\pi(s|a)\sum\limits_{s,r}P_{ss'}^{a}[r+\gamma V^{\pi}(s')]
\end{equation}

It gives the expected return for the agent when starting at state $s$ and following policy $\pi$. $r$ is the reward received for taking action $a$ while in state $s$ and transitioning to state $s'$. $\gamma \in [0,1]$ is a discount factor and determines the present value of future rewards. If $\gamma=0$, the agent is ``myopic" and only concerned with the immediate rewards. If $\gamma=1$, the agent strives towards long-term rewards. Equation \ref{eq:2_bellman} is known as the Bellman equation and it expresses the relationship between the value of a state and the value of its successor states \citep{Sutton2014}.

The approach to finding the solution of the MDP, i.e., determining the optimal policy that returns the largest expected reward by any policy, depends on whether the probability transitions $P_{ss'}^{a}$ and the reward function $r$, i.e., dynamics of the system, are known. Focusing on the case where these are unknown, the optimal policy must be estimated by the agent interacting with the environment. There are two available approaches at this point, either learning the model and then applying a planning procedure, or learning the optimal action for each state without explicitly learning the transition probabilities between the states \citep{Vazquez-Canteli2019}. The major advantage of this second method is that it is potentially a model-free control approach avoiding the need to create non-generalisable models.

\begin{figure}[!htbp]
\centering
\includegraphics[clip, trim=6cm 6.5cm 6cm 4.5cm]{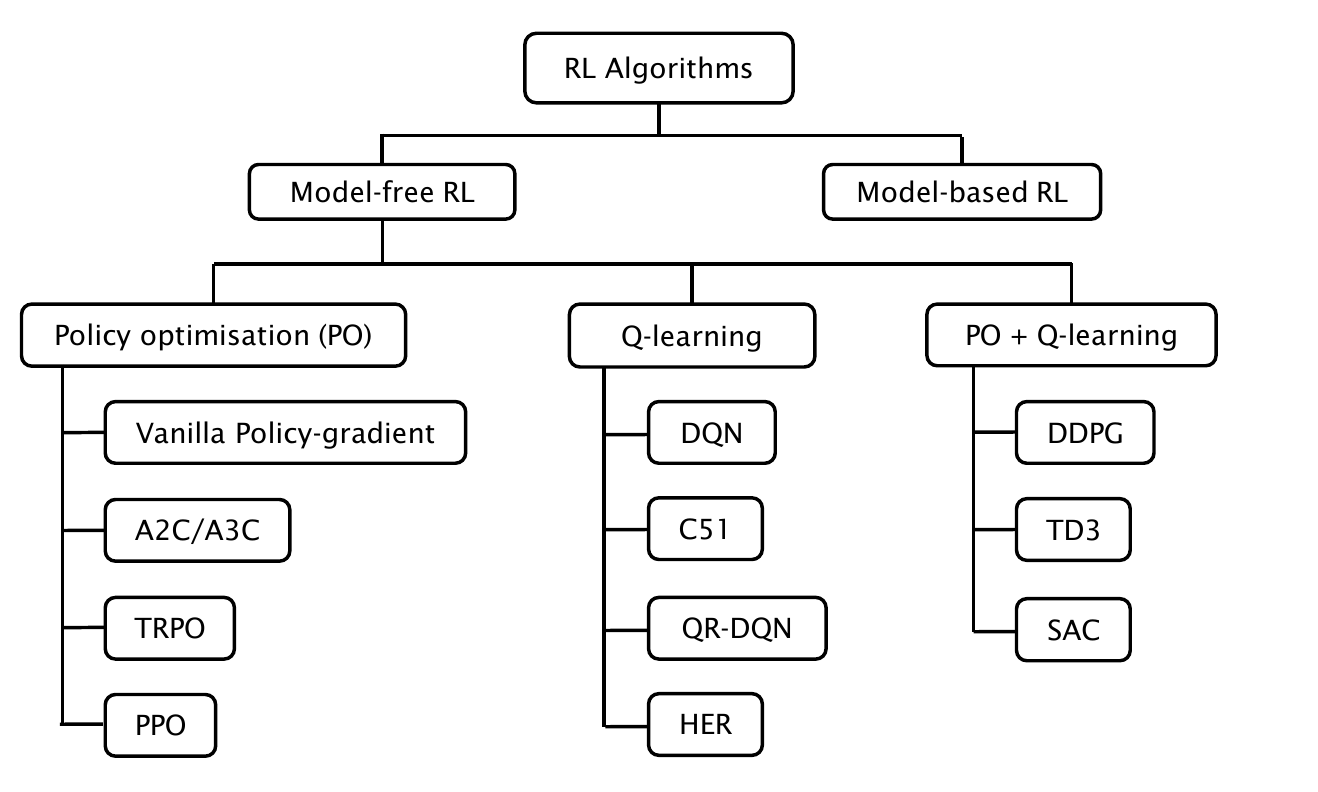}
\caption{Class of model-free RL algorithms (adapted from \citep{OpenAI2018})}
\label{fig:2_RL_algorithms}
\end{figure}

The model-free reinforcement learning algorithms can be divided into three main classes: (i) Policy optimisation, (ii) Q-learning, and (iii) a combination of policy optimisation and Q-learning \citep{Azuatalam2020} (see Figure \ref{fig:2_RL_algorithms}). Policy optimisation approaches employ an approximator $V_{\phi}(s)$ for the value function and this is used to generate the policy. Q-learning, where the state-action value function is learned, is an established RL method and commonly found in literature \citep{Costanzo2016}. Often, a table is used to represent the transitions in which the state-action values, or Q-values, are stored.

In RL, learning approaches can be categorised as either on-policy or off-policy. Off-policy algorithms do not need to follow a specific policy to update the Q-values, and this allows them to learn from historical data where actions were performed where any policy may have been applied. Another concept of importance is the trade-off between exploration and exploitation. To achieve convergence of the Q-values, it is necessary to explore actions for which the Q-values are not necessarily the highest yet, however at the cost of greater rewards. Exploration is trying actions that at a given iteration seem to be sub-optimal and exploitation is performing actions that seem to lead to the optimal policy. This trade-off is known as action-selection and there are several methods to deal with this problem \citep{Vazquez-Canteli2019}. The most widely used method is the $\epsilon$-greedy policy which consists of taking actions with the greatest Q-value with probability $(1-\epsilon)$, and selecting a random action with probability $\epsilon$. The soft-max action selection is also commonly used and the probability ($\rho$) of choosing an action uses a Boltzmann distribution with a temperature parameter $\alpha$ and is related to its Q-value as follows ($n$ is the number of possible actions):

\begin{myequation}
\label{eq:2_softmax}
  \rho_{(a_i|s)} = \frac{e^{\frac{Q(s,a_i)}{\alpha}}}{\sum\limits_{b=1}^n e^{\frac{Q(s,a_b)}{\alpha}}}
\end{myequation}

With large state-action spaces or where the states and actions are continuous, the speed of convergence is severely reduced due to the curse of dimensionality. The Q-table can be replaced by a function estimator, e.g., through the use of Deep Neural Networks/Deep Q Networks (DNN/DQN) resulting in Deep Reinforcement Learning (DRL). DNN have gained particular popularity due to their ability to build an effective representation of the problem through their hidden layer structure. Here, the neural network relates the value estimates and state-action pairs \citep{Wei2017}. Model-free deep RL algorithms have been applied in a range of challenging domains from games \citep{Mnih2013, Silver2017} to building energy management \citep{Brandi2020, Azuatalam2020, Mocanu2019}. One of the challenges often faced is sample efficiency with a large number, possibly ranging in the millions, of steps of data collection required. The on-policy nature of some of the policy optimisation approaches which often require new samples to be collected for each gradient step become computationally very expensive. Off-policy algorithms are able to reuse past experience and while not directly feasible with conventional policy gradient methods, are suitable for use with Q-learning based approaches \citep{Haarnoja2018}. However, the combination of off-policy learning and highly non-linear function approximation with DNNs presents stability and convergence challenges and hence the third class of model-free algorithms using a combination of policy optimisation and Q-learning have arisen.

\citet{Wang2020} provide an overview of reinforcement learning applied to building controls focusing on both the opportunities and challenges. Challenges identified by the authors include the training process being time-consuming and data-demanding, security of the cyber system, scalability to the building stock using transfer learning and finally the need for data-rich open source virtual test-beds for benchmarking. For an overview more focused on DR applications, the reader is referred to the review conducted by \citet{Vazquez-Canteli2019}.

The nature of the DR engineering problem is such that it features continuous state and action spaces (e.g., cooling setpoints, battery charge/discharge rates). Some of the most significant progress in the field of DRL has been in problems with discrete and low-dimensional action spaces, e.g., the use of DQN in Atari games \citep{Mnih2013} and AlphaGo Zero achieving superhuman performance starting tabula rasa \citep{Silver2017}. Whilst discretising the action space is one option to allow these DRL approaches to be adapted for continuous action environments, this poses challenges such as the curse of dimensionality \citep{Lillicrap2016}. This is especially an issue in situations where fine control of actions is required.

This current research utilises the soft actor-critic (SAC) DRL algorithm, an off-policy maximum entropy actor-critic algorithm, as first proposed by \citet{Haarnoja2018}, which is able to handle continuous action spaces. The authors suggest that the SAC algorithm provides for both sample-efficient learning and stability and hence extends readily to complex, high-dimensional tasks. They found the SAC algorithm showed substantial improvement in both performance and sample efficiency over both off-policy and on-policy prior methods. There have been a few recent examples where the SAC algorithm has been applied to the building energy management problem. \citet{Kathirgamanathan2020a} and \citet{Pinto2020} used the algorithm to optimally control a cluster of buildings using the CityLearn environment \cite{Vazquez-Canteli2019}. The CityLearn environment is an OpenAI environment which allows the control of domestic hot water and chilled water storage in a district environment. This environment was also used by \citet{Dhamankar2020} who compared three classes of multi-agent RL algorithms for demand response and coordination of a district. Note that for this environment, the conditioning loads were precomputed for a given set of climates. Whilst limited applications exist utilising the SAC algorithm for such DR problems, the algorithm has not been tested with a detailed simulation environment such as an EnergyPlus based one (nor a real-life implementation) to the best knowledge of the authors. The SAC algorithm features a number of different hyperparameters which control the agent learning behaviour and the sensitivity of these parameters to learning an optimal control policy has not been thoroughly investigated. Finally, the adaptability of the algorithm to deployment for changing or different environments is also a research gap. These research gaps lead to the following motivation and contributions of this research work which are highlighted next.

\subsection{Motivation and Structure of this Work}
\label{sec_motivation}

This research is motivated by the potential of model-free DRL techniques such as SAC for automated DR and avoiding the need to develop building-specific models. The SAC algorithm is particularly relevant given its ability to handle continuous action spaces as is found in the case of optimal control of HVAC systems. To date, the SAC DRL algorithm has been limited to simple environments (e.g., precomputed cooling loads) or non building energy management problems. There is the need to test the SAC DRL algorithm in a detailed high-fidelity simulation environment in lieu of a real-life implementation for the building energy flexibility problem. For such environments, the learning performance of the algorithm as well as its robustness is unknown. The application of the SAC DRL algorithm to a high-fidelity EnergyPlus based simulation environment is one of the contributions of this research. The state space choice and hyperparameter selection influence the design of the agent and ultimately the performance of the controller in deployment. This research contributes by providing a sensitivity study of both these parameters with regards to achieving the best control performance. Finally, the question of robustness and scalability of the SAC controller over a heterogeneous building stock is of interest and to address this, the ability of the agent to work under different climatic conditions is investigated. 

The paper is organised as follows: Section \ref{sec_SAC} provides essential background to the concept and formulation of the SAC algorithm. Section \ref{sec_methods} presents a detailed description of the environment developed to test the controller, and the methods used to design, train, tune and test the DRL controller. Section \ref{sec_results} provides the results and discussions of the key findings with focus on the training process, hyperparameter training, influence of state space, deployment of the agent and finally robustness of the DRL agent. Section \ref{sec_conclusions} gives the conclusions and summarises potential future research directions to enable and enhance the further use of the SAC DRL technique for real-life energy flexibility applications.

\section{Soft Actor Critic Deep Reinforcement Learning: Concept and Formulation}
\label{sec_SAC}

At their core, actor-critic methods are a type of policy gradient methods which have separate memory structures to explicitly represent the policy \citep{Sutton2014}. The policy structure is known as the actor and the estimated value function is known as the critic. The actor selects the actions whereas the critic evaluates the actions made by the actor. The reader is referred to Sutton \citep{Sutton2014} for a more detailed explanation of actor-critic methods. The soft actor-critic (SAC) algorithm, an off-policy maximum entropy actor-critic algorithm, as first proposed by \citet{Haarnoja2018} is used in this research. The fundamentals of this particular algorithm are presented next.

Where the SAC differs from traditional actor-critics is that SAC maximises the information entropy of state apart from the conventional cumulative rewards. Standard RL maximises the expected sum of rewards as outlined in Equation \ref{eq:2}.

\begin{equation}
\label{eq:2}
  J(\pi) = \sum\limits_{t}\EX_{(s_t,a_t)~\rho_\pi}[r(s_t,a_t)]
\end{equation}
SAC, however, favours stochastic policies and it does this by modifying the objective function with an additional term of the expected entropy ($\mathcal{H}$) of the policy:

\begin{equation}
\label{eq:entropy_opt}
  J(\pi) = \sum\limits_{t}\EX_{(s_t,a_t)~\rho_\pi}[r(s_t,a_t) + \alpha \mathcal{H} (\pi(\cdot|s_t))]
\end{equation}

Here $\alpha$ and $\mathcal{H} (\pi(\cdot|s_t))$ is the trade-off between entropy and reward. The advantage of entropy maximisation is that it can lead to policies that can explore more and are able to capture multiple modes of near-optimal strategies \cite{Haarnoja2018}. Increasing entropy can also prevent the policy from prematurely converging to a bad local optimum. This objective can be extended to infinite horizon problems by introducing the discount factor $\gamma$ and as derivation of this objective is more involved, the reader is referred to the plenary text of \cite{Haarnoja2018}. At test time, stochasticity is removed and the mean action is used instead of a sample from the distribution. 

A DNN is used to model the mean of the log of the standard deviation of the policy. During the policy improvement step, the policy distribution is updated towards the soft-max distribution for the current Q function by minimising the Kullback-Leibler divergence \cite{Kullack1951}. The algorithm makes use of two soft Q-functions (DNNs) to mitigate positive bias in the policy update step. These are trained off-line using data from the replay buffer collected during deployment. The minimum value from the two soft Q-functions is used in the gradient descent. Target networks are also employed using the exponential moving average of the soft Q-function weights to smooth out effects of noise in the sampled data. The SAC algorithm is summarised in Algorithm \ref{alg:SAC} with the full detail left for \cite{Haarnoja2018}.

The SAC DRL agent was developed in Python and using the PyTorch library \citep{paszke2017automatic} based off the implementation of Pranjal Tandon (\url{https://github.com/pranz24/pytorch-soft-actor-critic}) created for a different RL environment (MuJoCo). Modifications were required to make the agent compatible with the environment developed (and as outlined in Section \ref{sec_env}). PyTorch is an open source machine learning library providing tensor computation with strong GPU acceleration and DNN which are used to construct the SAC DRL.

\begin{center}
\begin{algorithm}[H]
\label{alg:SAC}
\SetAlgoLined
\KwResult{Optimised actor and critic DNNs}
 initialise policy, two soft Q and two target soft Q DNNs\;
 initialise experience replay buffer with random policy samples\;
 \For{each episode}{
  \For{each step}{
   sample actions from policy\;
   sample transition from the environment\;
   store the transition in the replay buffer\;
  }
  \For{each gradient update step}{
   update the soft Q DNN weights\;
   update the policy DNN weights\;
   update the target DNN weights\;
  }
 }
 \caption{SAC algorithm per \cite{Haarnoja2018} and \cite{Anderlini2020}}
\end{algorithm}
\end{center}

\section{Methods}
\label{sec_methods}

\subsection{Environment}
\label{sec_env}

\begin{figure}[!htbp]
\centering
\includegraphics[scale=0.6, clip, trim=1cm 2cm 0cm 2.8cm]{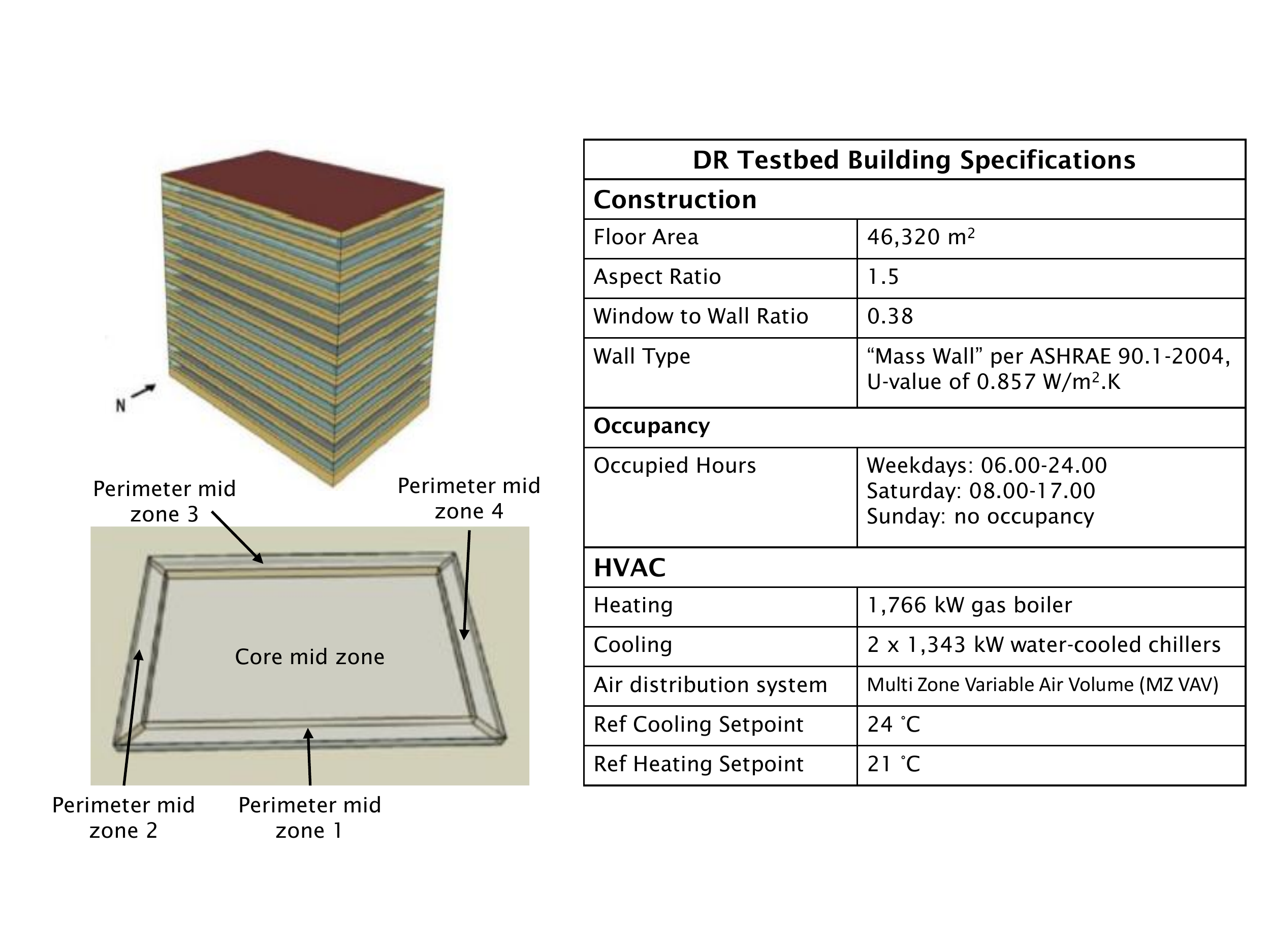}
\caption{Top left: 3D view of `Large Office' building, bottom left: zone plan of `Core Mid' zone, right: `Large Office' building specifications}
\label{fig:3_virtual_building_details}
\end{figure}

As a representative commercial building, a US-DOE (United States Department of Energy) commercial building archetype model was chosen \citep{Deru2011}. They are provided in the form of EnergyPlus input files. The DOE states that the archetype models represent approximately two-thirds of the commercial building stock in the USA \citep{Deru2011}. In this body of research, the `Large Office' reference building model was chosen as it represented the largest building by floor area. The version with ``new construction" was selected in particular with this building complying with the minimum building envelope and thermal properties of ASHRAE Standard 90.1-2004. Based on this standard, the building has a typical U-value of 0.857 W/m$^2$.K. This building has a floor area of 46,320 m$^2$ over 12 floors. The building operates from 6.00 am to midnight on weekdays and 6.00 am to 5.00 pm on Saturdays (there is no occupancy on Sundays). This information is summarised in Figure \ref{fig:3_virtual_building_details} together with a 3D view and a zone plan of the `core mid' zone that is considered in this research. The building is assumed to be located in Rome, Italy for the initial training of the RL agent (the issue of climate is considered later).

In terms of HVAC systems, the building has a gas boiler for heating, two water-cooled chillers for cooling and a multi-zone variable air volume systems for air distribution. Note that these systems are autosized using EnergyPlus based on the climate used for the dataset. The archetype model was modified through the addition of a thermal energy storage (TES) tank (100 m$^3$ volume (chilled)) to add a further flexibility source to the building. The chillers are operated in parallel configuration; the primary chiller directly meets the cooling load whereas the secondary chiller is used to charge the TES. The reference strategy is for the secondary chiller to charge the TES during the unoccupied hours and the TES discharges during the day, thereby reducing the primary chiller cooling load. 

This research focuses on the building `core mid' zone treating this zone as representative of the entire building. This is justified as this zone represents the majority of the zonal temperatures, being the largest zone per floor and representing 10 of the 12 floors through symmetry properties of the simulation (the EnergyPlus model only simulates one of these 10 `middle' floors and assumes the other nine floors are identical). 

In the absence of a real building, the same EnergyPlus model described in this section, acting as the surrogate model, is used to test the RL control in a closed-loop simulation through the use of co-simulation. This is necessary as EnergyPlus, whilst being a detailed physics based simulation engine, lacks the capabilities for implementation of more advanced control strategies and only allows manually coded rule-based control strategies. In most building automation systems (BAS), a supervisory control and data acquisition (SCADA) system is employed by operators to manage the various systems present in a building \citep{Jain2018a}. These SCADA systems interface directly with building sensors and low-level controllers through open source protocols such as BACNet or open platform communications (OPC or OLE for Process Control) \citep{InternataionalElectrotechnicalCommission2020}. To ensure compatibility of the developed data-driven controller with such SCADA systems, an external library is used to connect EnergyPlus using the OPC interface. In this case, inputs and outputs from EnergyPlus are represented using OPC tag structures hence ensuring the DRL control is transferable to a real building featuring a SCADA system. The PyEp python module is used for communication between EnergyPlus and Python (where the DRL control is implemented) through the use of an OPC bridge. The PyEp based controllers use the OpenOPC library to connect to an OPC server. The reader is referred to \citet{Jain2018a} for further information on PyEp and the EnergyPlus-OPC bridge.

\begin{figure}[!ht]
\centering
\includegraphics[scale=0.7, clip, trim=2cm 0cm 1cm 1cm]{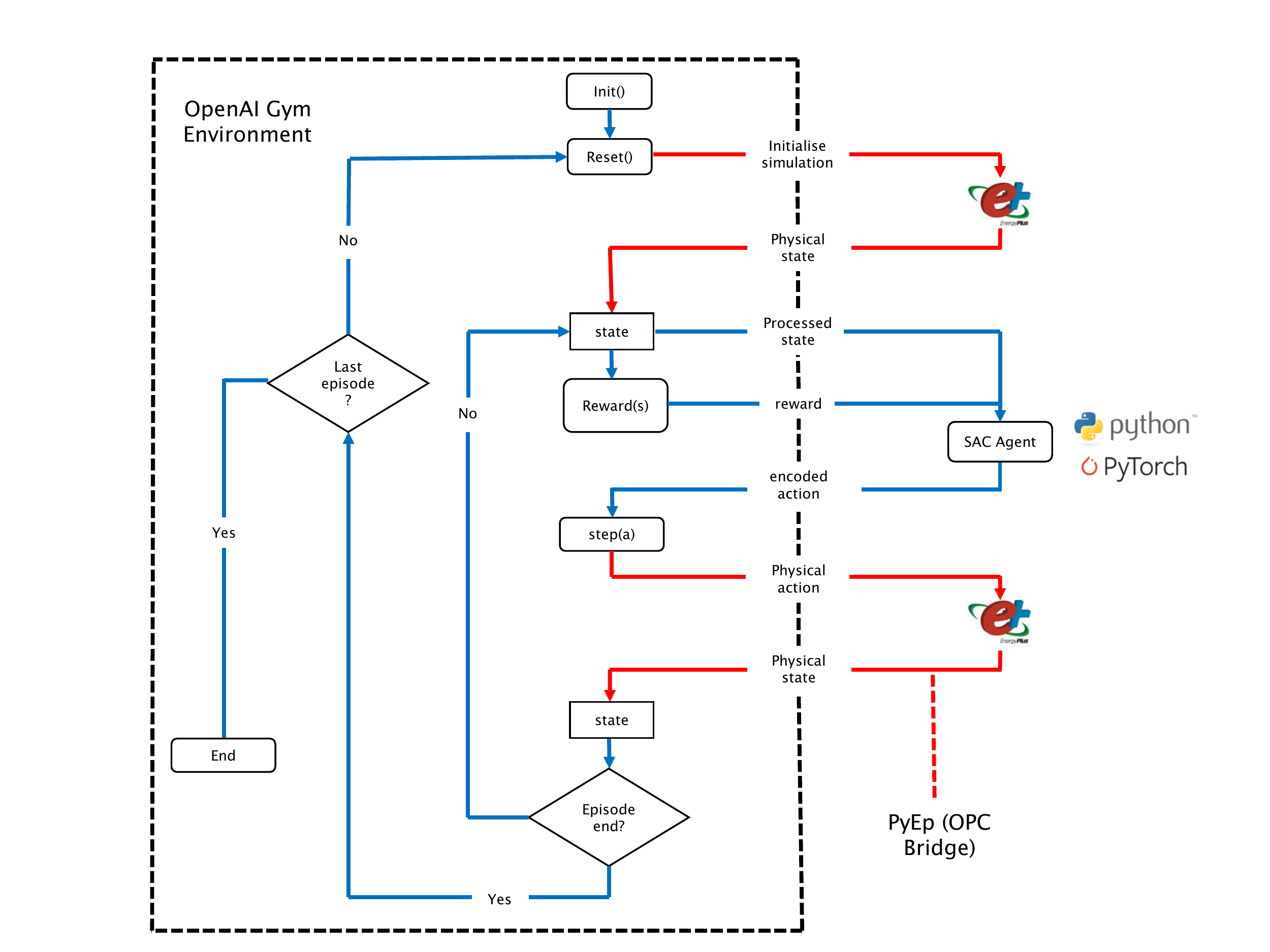}
\caption{Architecture of simulation environment for RL control in HVAC systems (modified from \citep{Brandi2020})}
\label{fig:RL_architecture}
\end{figure}

The detailed information flows in this architecture are described in further detail in Figure \ref{fig:RL_architecture}. The simulation starts with the initialisation (\emph{init()} function) of the OpenAI Gym environment. The \emph{reset()} function is called at the beginning of each episode (one simulation) where it initialises the EnergyPlus simulation process performing the simulation warm-up and returns the initial state of the building. The states returned by EnergyPlus need to be processed before being passed on to the DRL agent. Based on the input, the DRL agent selects a possible action and passes it to the \emph{step()} function which converts the passed value to a physical control action. The control variable is the cooling set point and this value is passed onto EnergyPlus as a schedule value using the PyEP OPC bridge. Once this happens, one further simulation timestep is completed in EnergyPlus and the new state of the building is returned through the OPC bridge. These new state values are used to determine the reward for the agent based on its actions. This process continues until the end of an episode is complete and many such episodes can be run to train the agent over a number of different histories or trajectories. A simulation time step of 15 minutes is used as an appropriate compromise between modelling the fast-dynamics and the data-burden from simulating short time steps over large prediction horizons \cite{Sturzenegger2016}. 

Other authors have implemented similar environments to test their RL agents as in the work of \citep{Brandi2020} and \citep{Vazquez-Canteli2018}. These implementations were, however, either for different RL agents (e.g., agents only operable in discrete action/state spaces) or different detailed simulation software (e.g., CitySim). It was noted that there is a lack of well established environments to allow co-simulation with EnergyPlus in the fashion described earlier and this is one of the challenges with implementing and testing such advanced control techniques as RL. This lack of suitable environments also provides a challenge for benchmarking different RL algorithms over different buildings \citep{Vazquez-Canteli2019}. Addressing these challenges are one of the contributions of this research.

\subsection{State Space}

For RL, the question of state space selection is considered to be a key part of the development of the DRL controller. The RL problem is initially framed such that the DRL agent is required to select the value of the cooling set point for the variable air volume system feeding the `core mid' zone (see Figure \ref{fig:3_virtual_building_details}). A range of 21 $^{\circ}C$ to 28 $^{\circ}C$ is set as the allowed range with the agent outputting normalised values between the range of 0 and 1. 

The state is what the DRL agent observes for each control step. The `Large Office' archetype EnergyPlus model allows over 13,000 variables to be requested. In this research, three configurations of input variables representing different levels of detail and sizes of state space are tested as outlined in Table \ref{table:state_space_design} and summarised below: 
\begin{enumerate}
    \item Set I represents a minimal set featuring only the electricity price (based on real market data from Italy for 2017 \citep{GME}), current values (1$^{st}$ lag term) of the state variables that influence the reward function (corresponding to purchased electricity and zonal temperature), proxy variables representing the day of the week and the hour of the day and current ambient weather parameters. Note that all variables are easily acquired.
    \item Set II represents a slightly larger state space also featuring information representing the state of charge of the thermal energy storage tanks, a detailed building occupancy schedule variable (representing occupancy as a fraction between 0 and 1) and lag terms of the state variables.
    \item Set III represents a large state space also including electricity price and weather forecasts over a certain prediction horizon. All predicted variables are assumed to be perfect forecasts.
\end{enumerate}  

In all cases, the variables were selected aiming to ensure that they provide the agent with all the necessary information to predict the immediate future rewards (aiming to satisfy the Markov property of the RL problem) and that they are feasible to be collected in a real-world implementation. All variables were scaled to be in the (0,1) range according to min-max normalisation as commonly employed with DRL algorithms \citep{Zhang2019, Brandi2020}. Normalisation was performed based on min and max limits that would be expected in operation over the range of climates used (Table \ref{table:climates}) for the investigation into the robustness of the agent (see Section \ref{sec:robustness}).

\begin{table}[!ht]
\centering
\begin{tabular}{p{11cm} c c c} 
 \hline
 \textbf{State Variable} & \textbf{Set I} & \textbf{Set II} & \textbf{Set III} \\ [0.5ex]
  & \textbf{(min)} & \textbf{(med)} & \textbf{(large)} \\
 \hline
 Electricity price (\EUR/kWh) & \checkmark & \checkmark & \checkmark \\
 Electricity price (\EUR/kWh) (up to 4 hours ahead) & \xmark & \checkmark & \checkmark \\
 Total HVAC Demand (kW) & \checkmark & \checkmark & \checkmark \\
 Total Power Demand (kW) & \checkmark & \checkmark & \checkmark \\
 Core mid zone temperature ($^{\circ}C$) & \checkmark & \checkmark & \checkmark \\
 day of week (1-7) & \checkmark & \checkmark & \checkmark \\
 hour of the day (1-24) & \checkmark & \checkmark & \checkmark \\
 Outside drybulb temperature ($^{\circ}C$) & \checkmark & \checkmark & \checkmark \\
 Outside wetbulb temperature ($^{\circ}C$) & \checkmark & \checkmark & \checkmark \\
 Outside wind speed ($m/s$) & \checkmark & \checkmark & \checkmark \\
 Outside wind direction (degrees) & \checkmark & \checkmark & \checkmark \\
 Outdoor air relative humidity (\%) & \checkmark & \checkmark & \checkmark \\ [1ex] 
 Direct solar radiation ($W/m^2$) & \checkmark & \checkmark & \checkmark \\ [1ex] 
 Chilled water storage tank 1 temperature ($^{\circ}C$) & \xmark & \checkmark & \checkmark \\ [1ex] 
 Building occupancy schedule & \xmark & \checkmark & \checkmark \\ [1ex] 
 HVAC Demand lag terms (up to 1 hour lag) (kW) & \xmark & \checkmark & \checkmark \\
 Core mid zone temperature lag terms (up to 1 hour lag) ($^{\circ}C$) & \xmark & \checkmark & \checkmark \\
 Outside drybulb temperature ($^{\circ}C$) (up to 4 hours ahead) & \xmark & \xmark & \checkmark \\
 Outside wetbulb temperature ($^{\circ}C$) (up to 4 hours ahead) & \xmark & \xmark & \checkmark \\
 Outdoor air relative humidity (\%) (up to 4 hours ahead) & \xmark & \xmark & \checkmark \\ [1ex]
 Direct solar radiation ($W/m^2$) (up to 4 hours ahead) & \xmark & \xmark & \checkmark \\ [1ex] 
 \hline
\end{tabular}
\caption{State space configurations used for model-free case (Sets I, II and III)}
\label{table:state_space_design}
\end{table}

\subsection{Reward Function}
\label{sec:rewards}

The reward function ($r$) was derived in this study and is based on an automated price-based DR scheme with dynamic pricing. The reward function is split into two competing terms, a cost term reflecting the price of electricity purchased from the grid and a comfort related term (Equation \ref{eq:rl_reward}):

\begin{equation}
\label{eq:rl_reward}
  r = -\beta \times e_{HVAC} \times e(t) - \lambda \times |t_{zone} - t_{lim}|
\end{equation}
where:
\[
    t_{lim} = 
\begin{cases}
    t_{min},& \text{if } t_{zone} < t_{min} \\
    t_{max},& \text{if } t_{zone} > t_{max}\\
    t_{zone},              & \text{otherwise}
\end{cases}
\]

The cost term is proportional to the component of power consumption that arises from the HVAC equipment ($e_{HVAC}$ - that arising from the cooling needs of the building) multiplied by the price of electricity ($e(t)$). The comfort term is proportional to the deviation of zonal temperature outside of pre-specified and time-varying temperature bounds ($t_{min}, t_{max}$). Weighting parameters $\beta$ and $\lambda$ are introduced to weight the importance of the two terms of the reward function based on stakeholder requirements and make the the competing objectives comparable. $\beta$ is fixed to a value of $1\times10^{-5}$ and the value of $\lambda$ is treated as an additional hyperparameter as outlined in Section \ref{sec:hyperparameters}.

\subsection{Hyperparameters}
\label{sec:hyperparameters}

There are numerous hyperparameters that are found in the implementation of the SAC DRL algorithm which influence the behaviour of the agent and hence the overall performance of the controller when deployed. A sensitivity study was performed to analyse the impact of the most important hyperparameters (based on the insights from \citet{Haarnoja2018}, albeit applied to an arguably simpler environment) on the performance of the SAC DRL controller with respect to the control objectives described in the previous section. These are summarised in Table \ref{table:rl_hyperparameters_search}. For each configuration, three runs were performed (each with a different initialisation seed) and an average was computed over the runs to account for some of the stochasticity implicit with the SAC algorithm. The remaining hyperparameters were left unchanged for the sensitivity study as they were found to be less significant for influence on training performance and these are listed in Table \ref{table:rl_hyperparameters_fixed}.

Considering the hyperparameters that are part of the sensitivity study, the discount factor ($\gamma$) determines the importance of future rewards over immediate rewards. As the value of $\gamma$ approaches 1, future rewards are weighted just as significantly as immediate rewards whereas a value approaching 0 places little or no significance on future rewards leading to a myopic agent. The reward temperature parameter ($\alpha$) controls the stochasticity of the policy and is an important parameter to balance exploration and exploitation for the SAC algorithm. It determines the relative importance of the entropy term against the reward. The version of SAC implemented in this paper assumes a constant entropy regularisation coefficient ($\alpha$) over the course of training. $\lambda$ is the comfort weighting term introduced in Section \ref{sec:rewards}.

\begin{table}[!htbp]
  \centering
  \small
  \caption{Different hyperparameter configurations implemented in the training phase for SAC DRL applied to `Large Office' building}
  \label{table:rl_hyperparameters_search}
  \begin{tabular}{ccc}
    \toprule
    \textbf{Symbol} & \textbf{Description} & \textbf{Training Values}\\
    \midrule
    $\gamma$ & Discount factor & 0.99, 0.95, 0.9 \\
    $\alpha$ & Reward temperature parameter & 0.05, 0.2 \\
    $\lambda$ & Comfort weighting term & 100, 500, 1000 \\
  \bottomrule
\end{tabular}
\end{table}

Considering the hyperparameters that are fixed and are not a part of the sensitivity study, the learning rate was found to be relatively insensitive to the final learned policy with stable learning for a large range of values. The learning rate represents the amount that the neural network weights are updated during each step of training. The target smoothing coefficient ($\tau$) reflects the amount that the separate target value network (which slowly tracks the actual value function thereby improving stability) is updated. SAC uses an exponentially moving average with smoothing constant $\tau$. \citet{Haarnoja2018} found that the range of suitable values for $\tau$ to be relatively wide (between 0.0001 and 0.1). The replay buffer contains the previously sampled states and actions (experiences) and batches sampled from here are used to update the function approximators using the stochastic gradients. A Gaussian distribution is used to add stochasticity to the policy. Initially, network weights are updated every 64 steps (corresponding to one update per day). The neural network architecture for both the actor and critic networks has two hidden layers composed of 64 neurons.

\begin{table}[!htbp]
  \centering
  \small
  \caption{Hyperparameters for training for SAC DRL applied to `Large Office' building}
  \label{table:rl_hyperparameters_fixed}
  \begin{tabular}{cccc}
    \toprule
    \textbf{Symbol} & \textbf{Description} & \textbf{Training Value}\\
    \midrule
    - & Learning rate & \num{1e-3} \\
    $\tau$ & Target smoothing coefficient & \num{3e-3} \\
    - & Replay buffer size & \num{2e6} \\
    - & Minibatch size & 2048 \\
    - & Policy & Gaussian \\
    - & Update interval & 96 (once per day) \\
    - & Hidden layer size & 64 \\
  \bottomrule
\end{tabular}
\end{table}

\subsection{Robustness}
\label{sec:robustness}

DRL has a reputation for requiring large amounts of training data and hence experience in training before achieving suitable control policies \citep{Haarnoja2018, Wang2020}. Even simple learning problems may end up needing millions of sample points. Unless there is availability of a high-fidelity simulation model for training the DRL agent offline, large training periods consisting of many episodes is unsuitable for real-life implementations in a building. Long training periods may be infeasible in real-life applications due to the exploration of sub-optimal control strategies which may lead to thermal discomfort or additional costs in operating the associated energy systems. Another negative aspect of long training periods is associated with degradation of performance associated with charge cycles in active electric storage (batteries) \cite{Oldewurtel2013}. The first challenge addressed here in this section aims to achieve the same control performance (in terms of cost saved and thermal comfort achieved) with significantly lower training times.

One of the most highlighted features of model-free reinforcement learning is the ability to adapt to changing or different environments. Whilst studies have investigated the ability of DRL to adapt to different operating conditions (e.g., weather conditions, occupancy and set point changes \citep{Brandi2020}), this has not been investigated with the SAC DRL. As \citet{Wang2020} note, there is a very limited understanding currently of how agents trained on one building can be successfully deployed on another building (i.e., a new environment). With this in mind, an agent trained on one climate (Rome, Italy) is tested on the different climates listed in Table \ref{table:climates}. The locations selected represent the full range of climates per the ASHRAE climate zone definitions. The adaptability of the DRL agent for a different season of the same climate is also tested with the agent trained on the summer months being deployed for the transition month of September.

\begin{table}[!ht]
\centering
\begin{tabular}{c c c} 
 \hline
 \textbf{ASHRAE} & \textbf{Location} & \textbf{Weather File} \\ [0.5ex] 
 \textbf{Climate Zones} & & \\ [0.5ex]
 \hline
 0 & Darwin, Australia & AUS\_NT.Darwin.941200\_IWEC.epw \\ 
 1 & Miami, USA & USA\_FL\_Miami.Intl.AP.722020\_TMY3 \\
 2 & Cairo, Egpyt & EGY\_Cairo.Intl.Airport.623660\_ETMY.epw \\
 3 & Rome, Italy & ITA\_Rome.162420\_IWEC.epw \\
 4 & Vancouver, Canada & CAN\_BC\_Vancouver.718920\_CWEC.epw \\ [1ex] 
 5 & Dublin, Ireland & IRL\_Dublin.039690\_IWEC \\ [1ex] 
 6 & Datong, China & CHN\_Shanxi.Datong.534870\_CSWD.epw \\ [1ex] 
 7 & Tampere, Finland & FIN\_Tampere.029440\_IWEC.epw \\ [1ex] 
 8 & Fairbanks, USA & US\_AK\_Fairbanks.Intl.AP.702610\_TMY3.epw \\ [1ex] 
 \hline
\end{tabular}
\caption{Climate zones and locations used to test adaptability of SAC DRL control}
\label{table:climates}
\end{table}

\section{Results and Discussion}
\label{sec_results}

The RL framework outlined in Section \ref{sec_methods} was implemented to design, train, tune and deploy the SAC agent. The results are presented in this section describing the training procedure (Section \ref{sec_results_training}), influence of the various hyperparameters (Section \ref{sec_results_hyper}), influence of the state space design (Section \ref{sec_results_ss}), deployment of the selected agent (Section \ref{sec_results_deploy}) and the robustness of the agent to different training periods, climates and seasons (Section \ref{sec_results_robust}).
 
\subsection{Training of DRL Agent}
\label{sec_results_training}

Initially, the DRL agent is trained purely on spring/summer data to focus on the cooling behaviour. One training episode (or one simulation) is taken to be three months of simulation from the start of April to the end of June. As training is conducted offline on a simulation environment, many episodes can be simulated repeatedly to allow the agent to explore many different trajectories. 50 episodes were used for training initially (selected after some experimentation and being conservative to ensure as much learning was captured as possible) and the question of what is an adequate training period is addressed in Section \ref{sec_results_robust}. On average, one episode took 2 min to be simulated on the server machine. An entire training period of 50 episodes took on average 100 min to be simulated.

To facilitate training of the SAC DRL agent, it was found that increasing the time between successive control actions from 15 minutes to 1 hour was necessary. Without this, during early episodes of training when the agent is predominantly exploring, large and frequent changes in the control set point cause the agent difficulty in learning the dynamics and relationship between the set point and the power and zonal temperature evolution. This highlights the importance of understanding the time constant of a building when employing a training strategy for the DRL agent.

\subsection{Hyperparameter Training}
\label{sec_results_hyper}

\begin{figure}[!htbp]
\centering
\includegraphics{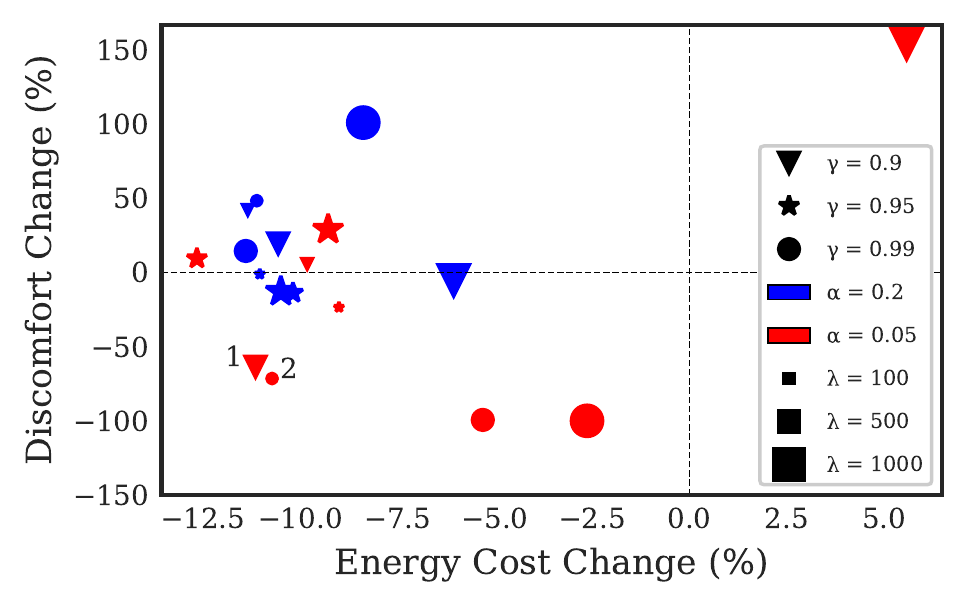}
\caption{SAC DRL agent control performance (x-axis: cost and y-axis:discomfort) compared against reference RBC for different combinations of hyperparameters as outlined in Table \ref{table:rl_hyperparameters_search}. The shape of the marker represents the discount factor ($\gamma$) of the experiment run, the colour represents the stochastic temperature term ($\alpha$) and finally the size of the marker represents the comfort weighting term ($\lambda$).}
\label{fig:5.8}
\end{figure}

The first results (Figure \ref{fig:5.8}) presented are for the different hyperparameter experiments carried out (per Table \ref{table:rl_hyperparameters_search}). The x-axis illustrates the percentage energy cost saving (defined in Equation \ref{eq:energy_spend_change}) compared to the reference rule-based control (RBC) for the test period when each agent is deployed (Section \ref{sec_results_deploy}). The RBC is simple preset time-based control of the cooling setpoint. The y-axis illustrates the percentage change in cumulative temperature constraint violations (defined in Equation \ref{eq:discomfort_change}) compared to the reference RBC. The shape of the marker represents the discount factor ($\gamma$) of the experiment run, the colour represents the stochastic temperature term ($\alpha$) and finally the size of the marker represents the comfort weighting term ($\lambda$). As this is a multi-objective optimisation problem with competing goals, a Pareto front can be obtained for different candidate solutions based on the weighting term. Note that the baseline deviation from comfort bounds for the RBC case is a very low number (0.60 hours) and so large percentage changes in discomfort are based on this low value. There appear to be two solutions that offer significant reductions in both energy cost and discomfort. One solution (labeled ``1" in Figure \ref{fig:5.8}) is with a $\gamma$ of 0.9 , $\alpha$ of 0.05 and comfort weighting of 500. The other solution (labelled ``2") is with a $\gamma$ of 0.99, $\alpha$ of 0.05 and a comfort weighting of 100. Based from these results, there are no significant obvious trends perhaps with the exception of a lower temperature parameter ($\alpha$) leading to better enforcement of comfort constraints. This highlights the necessity of hyperparameter tuning based on stakeholder requirements to obtain the Pareto front of solutions.

\begin{equation}
\label{eq:energy_spend_change}
  \Delta \textrm{ energy cost} = \frac{\textrm{energy cost}_{DRL} - \textrm{energy cost}_{RBC}}{\textrm{energy cost}_{RBC}} \times 100\%
\end{equation}

\begin{equation}
\label{eq:discomfort_change}
  \Delta \textrm{ discomfort} = \frac{\textrm{discomfort degree hours}_{DRL} - \textrm{discomfort degree hours}_{RBC}}{\textrm{discomfort degree hours}_{RBC}} \times 100\%
\end{equation}

\begin{figure}[!ht]
\centering
\includegraphics{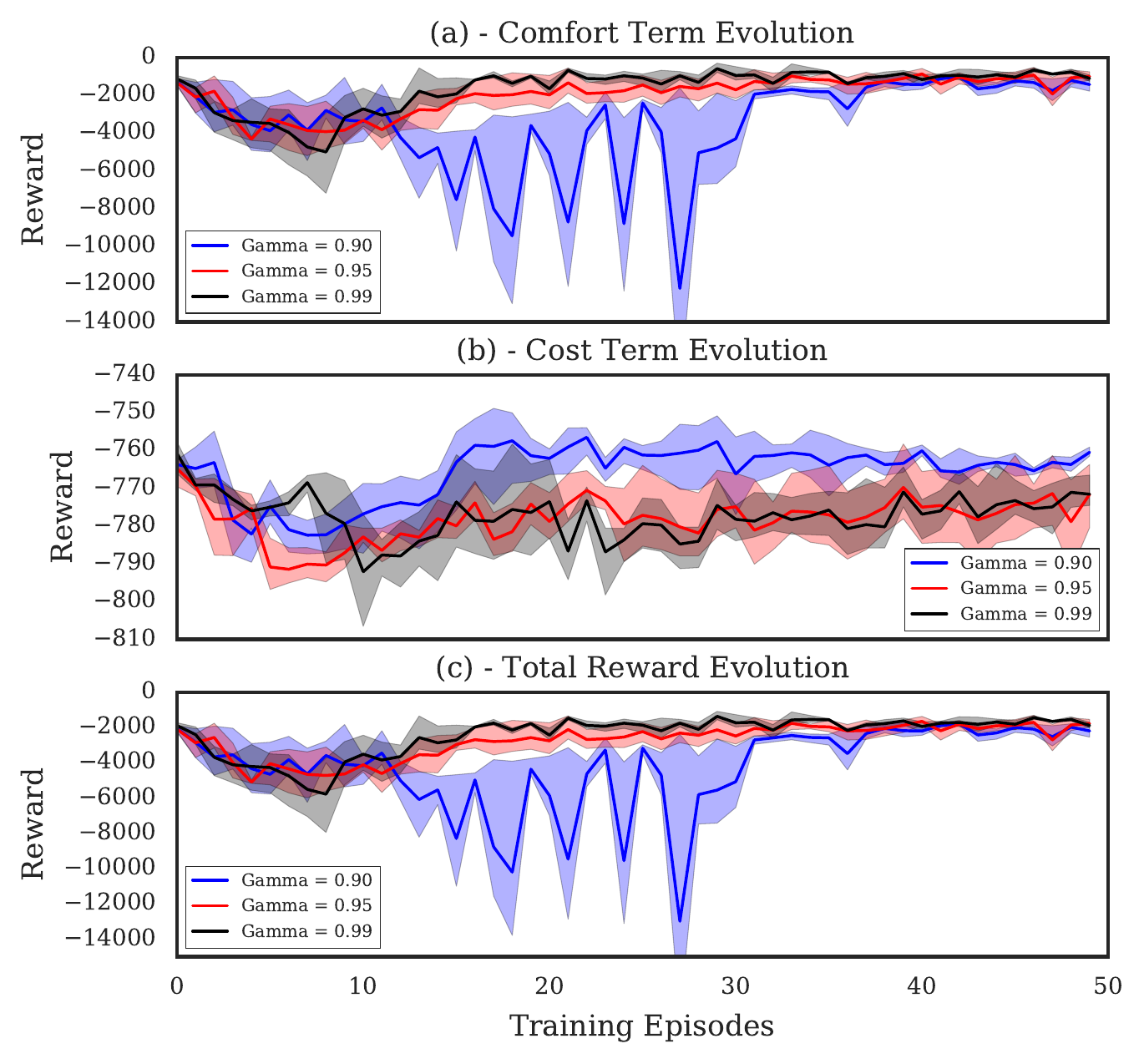}
\caption{Function of reward (Equation \ref{eq:rl_reward}) over training episodes for different discount (how myopic the agent is) factors ($\gamma$ = 0.9/0.95/0.99). The solid line represents the mean of the values whereas the filled area represents the bounds for the 25\% and 75\% quartiles.}
\label{fig:rl_training_discount_reward}
\end{figure}

\begin{figure}[!ht]
\centering
\includegraphics{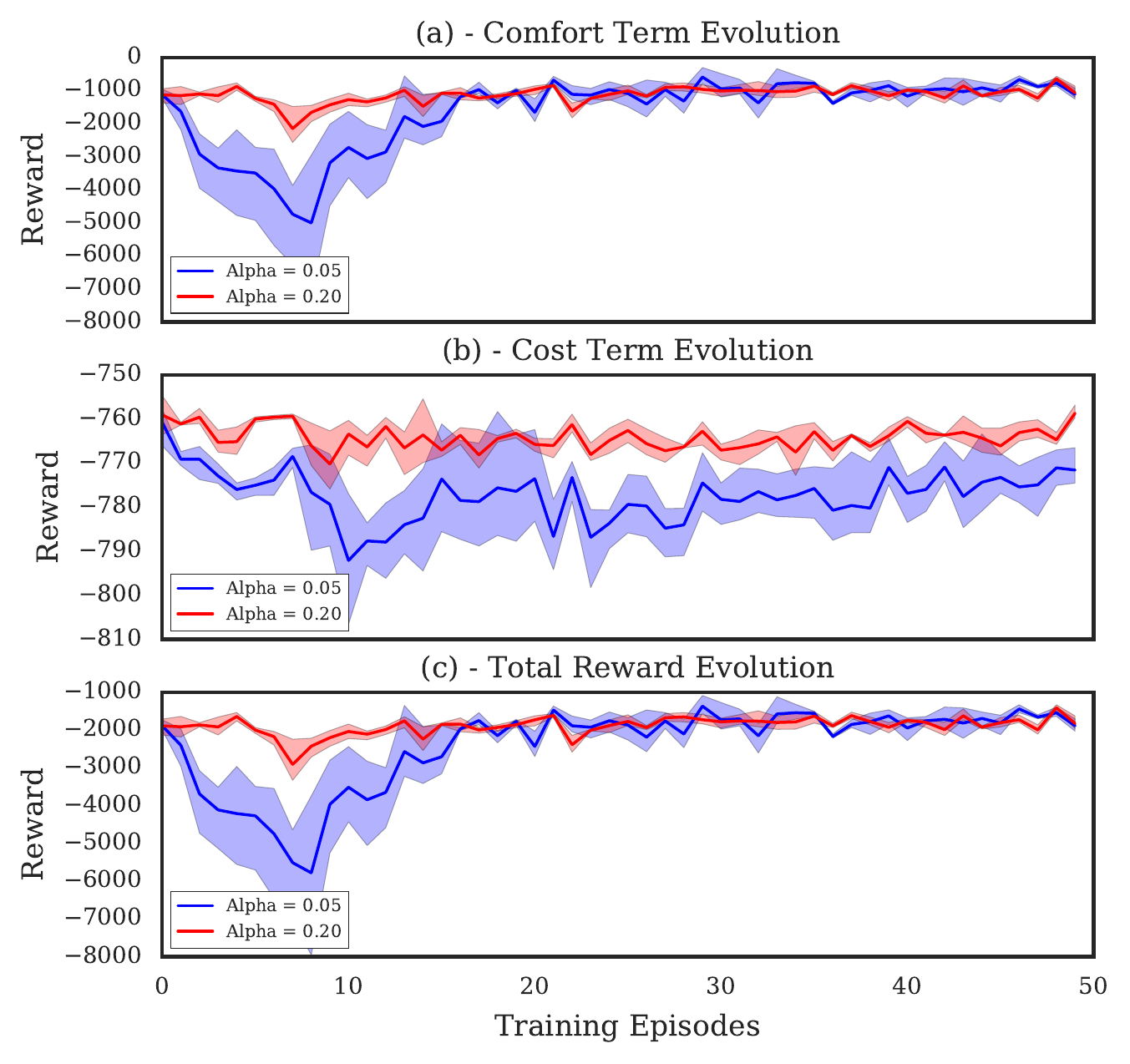}
\caption{Function of reward (Equation \ref{eq:rl_reward}) over training episodes for different temperature (to what level the agent explores the action space) parameters ($\alpha$ = 0.05/0.2). The solid line represents the mean of the values whereas the filled area represents the bounds for the 25\% and 75\% quartiles.}
\label{fig:rl_training_alpha_reward}
\end{figure}

The evolution of the reward terms (cost, discomfort and total - see Equation \ref{eq:rl_reward}) over the training episodes are illustrated in Figure \ref{fig:rl_training_discount_reward} for varying discount factors ($\gamma$) (and given a value of $\alpha$ of 0.05 and a comfort weighting of 500 from solution \#1). Note that the results presented are based on three runs for each configuration with different initialisation seeds to account for the stochasticity implicit in the SAC algorithm. The figure illustrates that the agent first learns to optimise thermal comfort (sacrificing the economic cost) and then aims to fine tune the cost term next. Figure \ref{fig:rl_training_alpha_reward} plots the evolution of the same reward terms for varying $\alpha$ terms (and constant discount rate of 0.99 and comfort weighting of 500). It shows that given a lower temperature term, a greater number of episodes are required to reach the optimal control policy (and indeed deviates from this first in exploration before learning from this experience). However, given 50 training episodes, both agents are able to reach similar performance by the end. These results reinforce the relative stability of the SAC algorithm to the different hyperparameters as reported by Haarnoja et al. and proves to be an advantage over more brittle algorithms for training \cite{Haarnoja2018}. 

\subsection{Influence of State Space}
\label{sec_results_ss}

\begin{figure}[!ht]
\centering
\includegraphics{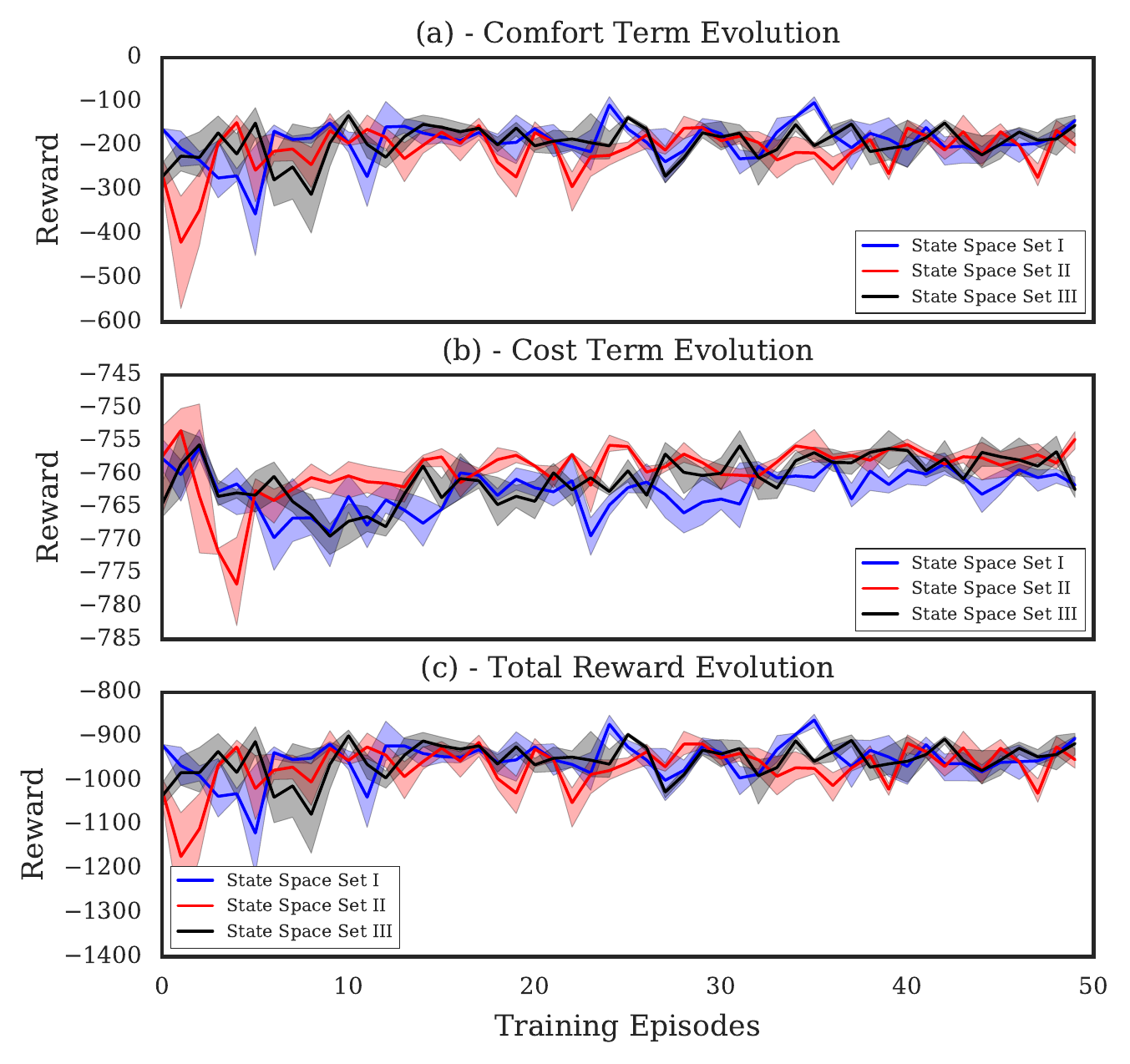}
\caption{Function of reward over training episodes for different state spaces (featuring differing number of input variables). The solid line represents the mean of the values whereas the filled area represents the bounds for the 25\% and 75\% quartiles.}
\label{fig:rl_training_statespace_reward}
\end{figure}

The next results (Figure \ref{fig:rl_training_statespace_reward}) are presented for the experiment with the different state spaces to investigate which variables the agent finds relevant and useful (given in Table \ref{table:state_space_design}). For these experiments, the parameters from optimal solution \#2 from the Pareto front is selected ($\gamma$ = 0.99, $\alpha$ = 0.05 and comfort weighting $\lambda$ = 100). Figure \ref{fig:rl_training_statespace_reward} plots the evolution of reward terms for the three different sets of states. The figure indicates that by the 50 training episodes, all agents essentially reach the same reward. When deployed, the performance of the different agents is summarised in Table \ref{table:SAC_state_space_performance}. The results show that the agents with the larger state spaces (i.e., set II and III) are able to make small improvements to the cost objective but it comes at a cost to discomfort for the largest state space. The medium state space (i.e., set II) is recommended following this analysis given the best combined cost and discomfort performance. However, whether this set would be robust to deployment in different boundary conditions or dynamic operating conditions should be further investigated. 

\begin{table}[!ht]
\centering
\caption{\protect\centering Comparison of performance of SAC DRL for different state spaces}
\label{table:SAC_state_space_performance}
\footnotesize
\begin{tabular}{c c c c}
 \hline
 \textbf{State Space} & \textbf{Energy Purchased} & \textbf{Energy Cost} & \textbf{Discomfort Degree-Hours} \\
  & \textbf{(MWh)} & \textbf{(\EUR)} & \\
 \hline
 State Space Set I & 105.91 & 5201 & 0.66  \\
 State Space Set II & 105.18 & 5150 & 0.14 \\
 State Space Set III & 106.18 & 5096 & 2.08 \\
 \hline
\end{tabular}
\end{table}

\subsection{Testing of DRL Agent}
\label{sec_results_deploy}

The final selected pre-trained agent ($\gamma$ = 0.99, comfort weighting = 100, $\alpha$ = 0.05 and state space set II) was run in evaluation mode (or test mode) for the same work week in July as for the model-based approach. As SAC converges to stochastic policies, for the best performance, it is recommended to make the final policy deterministic \citep{Haarnoja2018}. This is achieved by approximating the maximum a posteriori action by choosing the mean of the policy distribution in deployment mode \cite{Haarnoja2018}. This may degrade new learning and adapting to different environments in deployment. As the deployment period is initially set to one work week, new learning and adaptability is considered in Section \ref{sec_results_robust}.

\begin{figure}[!ht]
\centering
\includegraphics{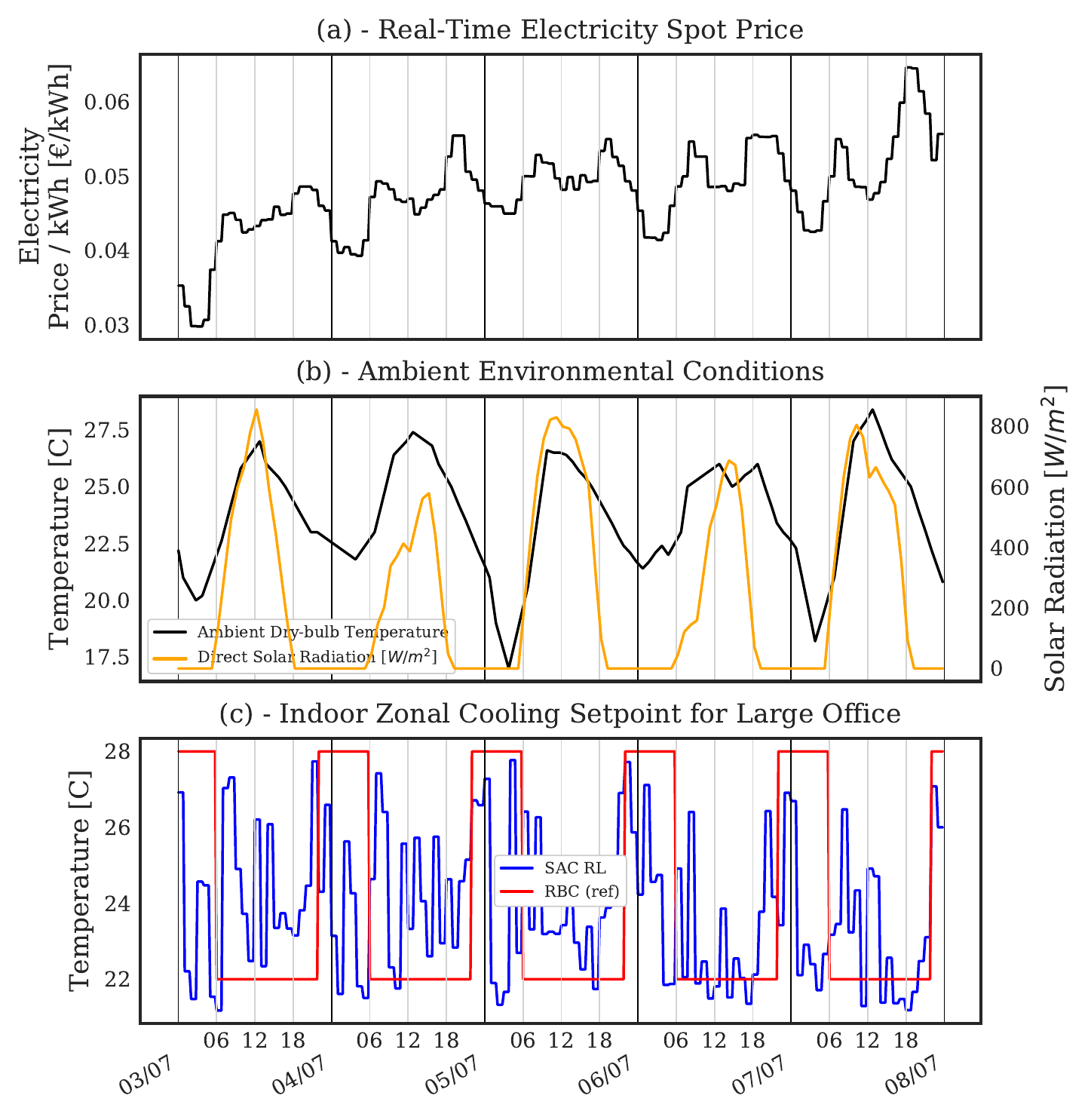}
\caption{SAC DRL results (compared to the baseline RBC) over work week in July (03/07-07/07) for test climate of Rome showing (a) real-time electricity spot price, (b) ambient environmental conditions and (c) actuation of control cooling setpoint}
\label{fig:rl_deployment_1}
\end{figure}

\begin{figure}[!ht]
\centering
\includegraphics{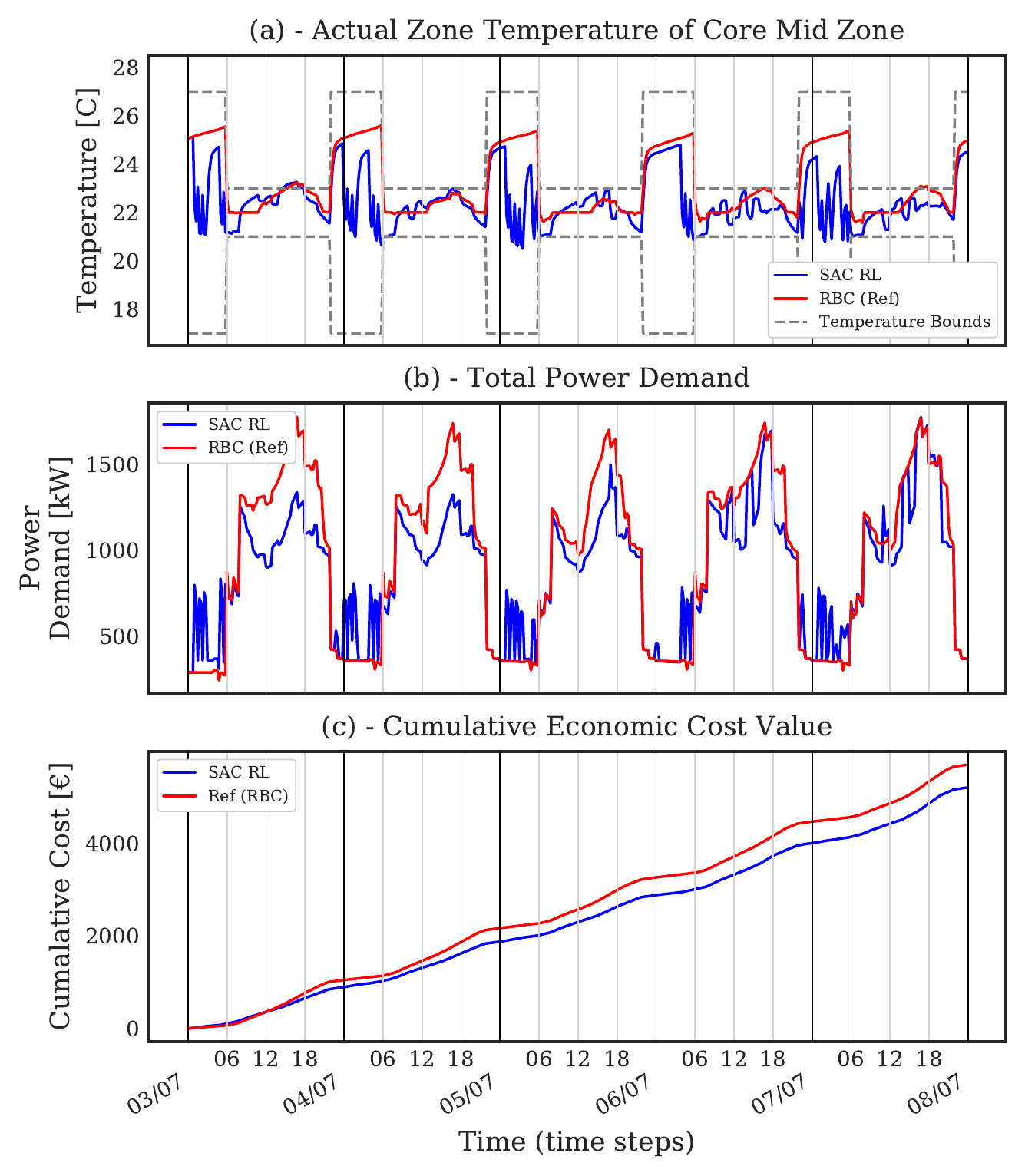}
\caption{SAC DRL results (compared to the baseline RBC) over work week in July (03/07-07/07) for test climate of Rome showing (a) actual zonal temperature, (b) building total power demand and (c) cumulative economic cost of control strategy}
\label{fig:rl_deployment_2}
\end{figure}

Figure \ref{fig:rl_deployment_1}(a) describes the real-time electricity price used as input by the controller which shows the trend of having lower prices during night hours (midnight to 06:00) and higher prices during the evening peak (17:00 to 20:00). Figure \ref{fig:rl_deployment_1}(b) illustrates the ambient external conditions. Figure \ref{fig:rl_deployment_1}(c) shows the control variable, in this case, the optimal scheduled indoor zonal cooling set point. It can be seen that the controller is able to pre-cool the building during the early morning taking advantage of lower real-time prices and hence requiring less cooling over the work day when higher prices are prevalent. This is also illustrated in Figure \ref{fig:rl_deployment_2}(b) showing the total power demand of both control strategies.

Figure \ref{fig:rl_deployment_2}(a) illustrates the zonal temperature evolution of the core mid zone for the deployment period. It is seen that the agent is able to respect the comfort constraints with just a few minor and short deviations outside the limits. Finally, Figure \ref{fig:rl_deployment_2}(c) plots the cumulative cost of electricity purchased for the SAC DRL agent and compares it with the RBC controller. The savings are detailed in Table \ref{table:SAC_performance}. The selected SAC DRL agent is able to realise a 8.8 \% saving in the amount of energy purchased, a 9.7 \% saving in the cost of energy purchased and 78.3 \% saving in discomfort hours. Note that there does seem to be some oscillatory control action during the night hours when the control is pre-cooling the building and this could lead to long term actuation fatigue. This is a current limitation of the work and minimising this requires further work, one possible solution being the addition of another reward term to account for the amount of actuation.

\begin{table}[!ht]
\centering
\caption{\protect\centering Comparison of performance of SAC DRL compared to reference RBC control over work week in July (03/07-07/07) for test climate of Rome for `Large Office' building}
\label{table:SAC_performance}
\footnotesize
\begin{tabular}{c c c c c c c}
 \hline
 \textbf{Control Type} & \multicolumn{2}{c}{\textbf{Energy Purchased}} & \multicolumn{2}{c}{\textbf{Energy Cost}} & \multicolumn{2}{c}{\textbf{Discomfort Degree-Hours}} \\
  & \textbf{(MWh)} & \textbf{(\% saving)} & \textbf{(\EUR)} & \textbf{(\% saving)} & \textbf{(-)} & \textbf{(\% saving)} \\
 \hline
 RBC (Ref) & 115.34 & - & 5702 & - & 0.60 & -  \\
 SAC DRL & 105.18 & 8.8 & 5150 & 9.7 & 0.13 & 78.3 \\
 \hline
\end{tabular}
\end{table}

Specific hyperparameters may be modified for deployment. The learning rate is one such parameter and controls how adaptable the controller should be to new operating conditions or climates. A lower learning rate (approaching 0) should be used where no significant changes in the operating conditions are expected from that used for training the agent. A higher learning rate should be used where the controller is expected to adapt to either a different building, different climate or changing operating conditions (e.g., occupancy patterns). This also presents risks as the controller may deviate from optimal performance given new training regimes.
 
\subsection{Robustness of DRL Agent}
\label{sec_results_robust}

This section presents results from an investigation into the robustness and scalability of the DRL agent. The investigation focuses on the length of training and agent update frequency required to achieve an optimal control policy (Section \ref{sec_results_robust_period}), adaptability when the agent is deployed to a different season (Section \ref{sec_results_robust_season}) and adaptability when the agent is deployed to different climates (Section \ref{sec_results_robust_climate}).

\subsubsection{Training Period and Season}
\label{sec_results_robust_period}

In the previous section, the SAC DRL was trained on 50 episodes (essentially 50 periods of training of 3 months (April - June) duration each). In a real-life implementation, if a high-fidelity surrogate training model is not available, this is not practical. The speed of learning was investigated through evaluating different training length periods (1 episode, 2 episodes, 5 episodes, 10 episodes and 50 episodes) for the algorithm. One of the hyperparameters that is very pertinent to the rate of learning is the network update parameter (Update interval parameter in Table \ref{table:rl_hyperparameters_fixed}, i.e., how often the gradient descents are performed on the networks). In the previous sections (Section \ref{sec_results_training} to \ref{sec_results_deploy}), during the development of the SAC DRL algorithm, the soft Q, policy and target DNN weights were updated every 96 steps (i.e., once per day). In this section, three different update frequencies are considered, 4 steps (every hour), 96 steps (every day) and 672 steps (once per week). All other hyperparameters are kept constant as per Table \ref{table:rl_hyperparameters_fixed} and the climate of Rome, Italy is assumed. A summary of the different runs is presented in Table \ref{table:6_rl_training_periods_search}.

\begin{table}[!htbp]
  \centering
  \small
  \caption{Different hyperparameter configurations implemented in the training phase}
  \label{table:6_rl_training_periods_search}
  \begin{tabular}{ccc}
    \toprule
    \textbf{Description} & \textbf{Training Values}\\
    \midrule
    Training Episodes & 1, 2, 5, 10, 50 \\
    Update Frequency (steps) & 4, 96, 672 \\
    Season & Summer \\
  \bottomrule
\end{tabular}
\end{table}

\begin{figure}[!htbp]
\centering
\includegraphics{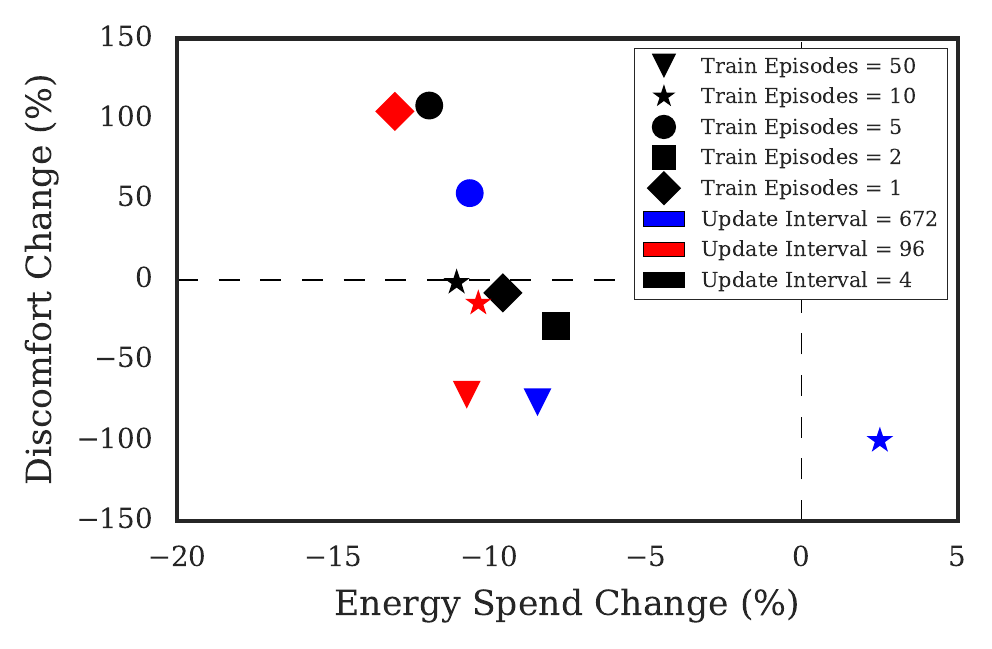}
\caption{SAC DRL agent control performance for different number of training episodes and update frequency for test work week of 03/07-07/07. The x-axis illustrates the percentage energy cost saving compared to the reference RBC (see Equation \ref{eq:energy_spend_change}) and the y-axis illustrates the percentage change in cumulative temperature constraint violations compared to the reference RBC (see Equation \ref{eq:discomfort_change})}
\label{fig:6_RL_training_period}
\end{figure}

Figure \ref{fig:6_RL_training_period} outlines the agent performance when evaluated on the same work week in July (03/07 - 07/07) for different combinations of training periods (number of episodes) and update frequencies. The x-axis illustrates the percentage energy cost saving compared to the reference RBC for the test period when each agent is deployed (Equation \ref{eq:energy_spend_change}). The y-axis illustrates the percentage change in cumulative temperature constraint violations compared to the reference RBC (Equation \ref{eq:discomfort_change}). The shape of the marker represents the training period of the experiment run and the colour represents the update frequency. Note that the baseline deviation from comfort bounds for the RBC case is a very low number (0.48 hours) and so large percentage changes in discomfort are based on this low value. Any run with a discomfort change of over 150\% is excluded from the figure (five excluded). The figure shows that it is possible to obtain comparable performance with only one episode of training (the black diamond in the figure and corresponding to three months of operation - hereby referred to as Agent X) with the update frequency needing to be at least once per hour.

Whilst one episode (three months) for training is a much lower training duration, unless there is a surrogate simulation or test site for conducting the training in, any online training on an occupied building must minimise any potential discomfort or sub-optimal energy performance or will be considered to be disruptive and potentially high-risk. With this in mind, the control performance of the agent during the training period is analysed next. The comparison of the controller performance (SAC DRL under training) with the reference RBC is presented in Table \ref{table:6_SAC_performance_training} for the case of Agent X. It can be seen that the agent is actually able to improve on the controller objectives (both energy cost and discomfort) even in the training period. This highlights the promising application of SAC DRL in a real-life online building with minimal disruption to thermal comfort and minimal cost in energy performance.

\begin{table}[!ht]
\centering
\caption{\protect\centering Comparison of performance of SAC DRL compared to reference RBC control during training period of April to June (one episode with update frequency every four steps)}
\label{table:6_SAC_performance_training}
\footnotesize
\begin{tabular}{c c c c c c c}
 \hline
 \textbf{Control Type} & \multicolumn{2}{c}{\textbf{Energy Purchased}} & \multicolumn{2}{c}{\textbf{Energy Cost}} & \multicolumn{2}{c}{\textbf{Discomfort Degree-Hours}} \\
  & \textbf{(MWh)} & \textbf{(\% saving)} & \textbf{(\EUR)} & \textbf{(\% saving)} & \textbf{(-)} & \textbf{(\% saving)} \\
 \hline
 RBC (Ref) & 1339.88 & - & 64049 & - & 1.47 & -  \\
 SAC DRL & 1333.65 & 0.5 & 63507 & 0.8 & 0.52 & 64.6 \\
 \hline
\end{tabular}
\end{table}

\subsubsection{Deployment to Different Season}
\label{sec_results_robust_season}

The SAC DRL is required to be able to optimise the control objectives under varying seasons and climatic conditions representing real-life seasonal operation. Whether an agent trained on one season may continue to have superior performance compared to RBC when deployed in a different season is investigated. In this case, the same agent trained on the summer season (Agent X, i.e., one episode of training corresponding to the three months from April to June, update frequency = four steps or once per hour) is deployed in the first work week of September (transition season for the climate of Rome, Italy). Table \ref{table:SAC_performance_transition} details the controller performance metrics comparing the SAC DRL controller with the reference RBC for the test period. In this case, for the test work week, the agent is able to still realise savings of 2.7\% for the energy consumption, 3.5\% for the energy cost and maintain thermal comfort levels. These results suggest that the SAC DRL controller is able to generalise to new seasons to a limited extent. To achieve optimal performance however, either tuning of the hyperparameters and weighting terms may be required and/or that the agent requires experience training on the new seasonal data.

\begin{table}[!ht]
\centering
\caption{\protect\centering Comparison of performance of SAC DRL compared to reference RBC control for agent trained on summer data tested over work week in September (04/09-08/09) for test climate of Rome}
\label{table:SAC_performance_transition}
\footnotesize
\begin{tabular}{c c c c c c c}
 \hline
 \textbf{Control Type} & \multicolumn{2}{c}{\textbf{Energy Purchased}} & \multicolumn{2}{c}{\textbf{Energy Cost}} & \multicolumn{2}{c}{\textbf{Discomfort Degree-Hours}} \\
  & \textbf{(MWh)} & \textbf{(\% saving)} & \textbf{(\EUR)} & \textbf{(\% saving)} & \textbf{(-)} & \textbf{(\% saving)} \\
 \hline
 RBC (Ref) & 106.89 & - & 5729 & - & 0.18 & -  \\
 SAC DRL & 104.38 & 2.3 & 5535 & 3.4 & 0.12 & 33.3 \\
 \hline
\end{tabular}
\end{table}

\subsubsection{Deployment to Different Climates}
\label{sec_results_robust_climate}

\begin{figure}[!htbp]
\centering
\includegraphics{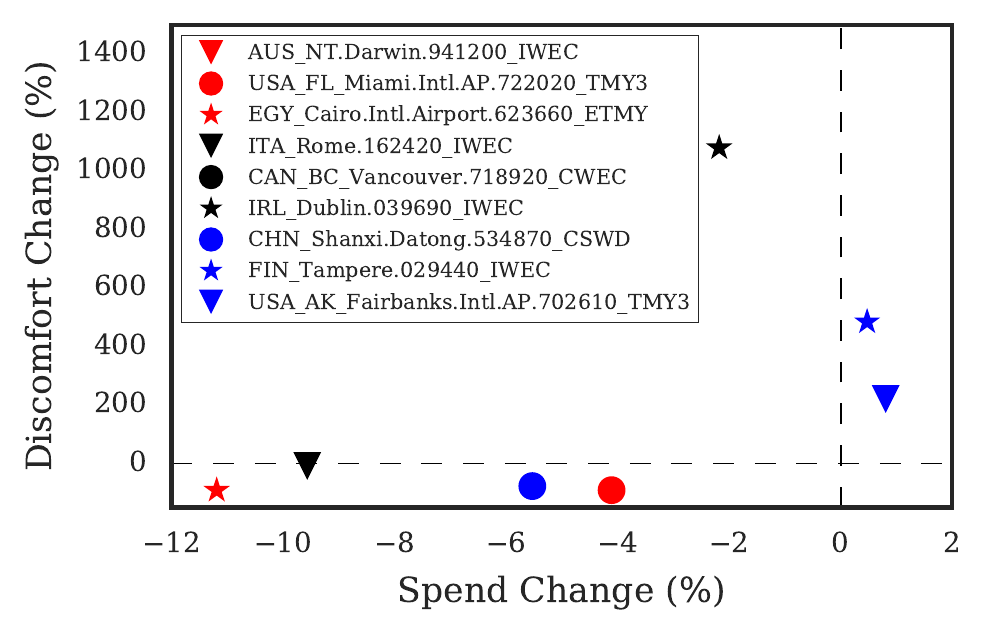}
\caption{SAC DRL agent trained on climate of Rome, Italy (ASHRAE climate zone 3) control performance (in terms of energy cost change (see Equation \ref{eq:energy_spend_change}) and discomfort degree-hours change (see Equation \ref{eq:discomfort_change}) compared to reference RBC) for deployment on other climates of Table \ref{table:climates}}
\label{fig:RL_training_climate_deploymentonly}
\end{figure}

Lastly, the adaptability of the SAC DRL for different climatic conditions is tested. The SAC DRL is deployed in different climates (Table \ref{table:climates}) than the one it was trained on. The SAC DRL (Agent X) trained on one episode (April to June) of the climate of Rome, Italy (ASHRAE climate zone 3) is deployed to the summer work week (03/07-07/07) period of the other eight climates. Note that sizing of the HVAC systems is performed according to the design day conditions of the relevant climate and site location for each of the simulations. Figure \ref{fig:RL_training_climate_deploymentonly} plots the Pareto front in this case. Note that the change in energy cost and change in discomfort hours (per Equations \ref{eq:energy_spend_change} and \ref{eq:discomfort_change}) are plotted with respect to the performance of the RBC controllers. The baseline discomfort for the RBC is a very low value for most climates which causes the large percentage changes in discomfort when comparisons are made. The SAC DRL agent is able to generalise very well, i.e., perform comparably, to the climate of Cairo, Egypt (ASHRAE climate zone 2) and even outperforms the climate on which the agent was initially trained on (Rome, Italy). The agent is also able to realise cost savings of 5.5 and 4.1 \% in the climates of Shanxi, China (climate zone 6) and Miami, USA (climate zone 1), respectively. For the other climates, the cost savings are not significant or the discomfort hours are increased significantly (e.g., in the case of the climates of Darwin and Vancouver). In these cases, the agent needs to be retrained on these climatic conditions perhaps with a sensitivity study required on the hyperparameters. Nevertheless, the deployment period in this study was only one work week which even given an update frequency (of the SAC DRL network) of once per hour, gives a limited window for the agent to adapt to the new environment. Longer deployments may yield different results and this is the subject of future work.

\section{Conclusions and Future Work}
\label{sec_conclusions}

A model-free deep reinforcement learning (DRL) `Soft Actor Critic' (SAC) approach for unlocking building energy flexibility for a commercial building was investigated in this research. The technique was evaluated in a surrogate simulation model of the DOE `large office' building through the use of co-simulation. This research provides the following contributions:
\begin{enumerate}
    \item Evidence of a model free DRL technique (which militates the need for building-specific models) which appears as a suitable candidate for automated building energy management and harnessing energy flexibility from the passive thermal mass.
    \item Consideration of the SAC DRL algorithm which is able to handle continuous action/state spaces and that has not been used in conjunction with an EnergyPlus based (high-fidelity) environment in previous literature.
    \item Consideration of the robustness of the algorithm through a sensitivity study of the hyperparameters.
    \item Investigation into the adaptability of the agent when deployed to different environments (climates).
\end{enumerate}

The SAC DRL agent was found to be able to learn an optimal control policy to activate building energy flexibility given a dynamic price signal. For a summer work week for the climate of Rome, Italy, the DRL agent was able to minimise energy costs (by 9.7\%) and maintain or even improve thermal comfort for occupants by respecting comfort constraints. This also highlights the possibility of this technique to take human feedback for occupant thermal comfort rather than temperature constraints. This was all possible without the need for training data-driven building and climate specific models. The technique has a minimal computational burden for online deployment suggesting its potential for harnessing energy flexibility at sub-hourly resolutions. A sensitivity study of the hyperparameters of the SAC algorithm was conducted showing the relatively robust nature of the technique to selection of hyperparameter values. The feature selection problem was also considered with three sets of state space considered in the design of the agent. The study showed that the agent was again relatively robust to the state space input and was able to outperform the rule-based controller (RBC) with a minimal state space consisting of readily available variables. The only variables required were the real-time electricity price, total HVAC power demand, total power demand, zonal air temperature, day of the week, hour of the day and ambient weather conditions.

The investigation into the robustness of the SAC DRL agent was undertaken in this research focusing on the length of training required and the ability of the agent to deploy to different climates and seasons. The SAC DRL requires minimal training sample points with the agent able to outperform the baseline reference RBC within one episode of training (3 months during summer). This is also possible without disruption to the building occupants in terms of thermal comfort or energy costs. The agent is not always perfectly transferable to different climatic conditions and either retraining and/or hyperparameter tuning is required in these cases. 

There are several limitations with this current study. Given the slower dynamics of a large building, the control timestep needed to be increased to a resolution of one hour for the agent to be able to learn the dynamics during training. It is unknown how the DRL agent performs in buildings with different dynamics to the `large office' building, especially smaller buildings with lower thermal mass and a greater proportion of the cooling load. The extension of this technique to a diverse set of buildings is left for future work. There are future plans to publish the environment as open source code in a way such that a user can easily interchange any of the DOE archetype models to further facilitate benchmarking of this particular algorithm. In terms of the adaptability and robustness study, one of the limitations was not deploying and testing the agents over a longer period which would allow the agent to dynamically learn and update to any new environment. The agent was only deployed for a period of one week which allows limited potential for updating the network weights and modifying the optimal control policy. Consideration of such dynamic deployment is also left for future work.


\section*{Acknowledgements}
\label{sec_ack}

The authors gratefully acknowledge that their contribution emanated from research supported by Science Foundation Ireland under the SFI Strategic Partnership Programme Grant Number SFI/15/SPP/E3125.


\renewcommand\nomgroup[1]{%
        \item[\bfseries
        \ifstrequal{#1}{A}{Acronyms}{%
        \ifstrequal{#1}{R}{Roman Symbols}{%
        \ifstrequal{#1}{G}{Greek Symbols}{%
        \ifstrequal{#1}{Sb}{Subscripts}{%
        \ifstrequal{#1}{Su}{Superscripts}{%
        \ifstrequal{#1}{X}{Other Symbols}{}}}}}}]%
    }

\makenomenclature
\printnomenclature

\Nomenclature[A]{ASHRAE}{American Society of Heating, Refrigerating and Air-Conditioning Engineers}
\Nomenclature[A]{BAS}{Building Automation System}
\Nomenclature[A]{BEMS}{Building Energy Management Systems}
\Nomenclature[A]{DSM}{Demand Side Management}
\Nomenclature[A]{DNN}{Deep Neural Network}
\Nomenclature[A]{DOE}{Department of Energy (US)}
\Nomenclature[A]{DQN}{Deep Q Networks}
\Nomenclature[A]{DR}{Demand Response}
\Nomenclature[A]{DRL}{Deep Reinforcement Learning}
\Nomenclature[A]{GPU}{Graphical Processing Unit}
\Nomenclature[A]{HVAC}{Heating Ventilation and Air Conditioning}
\Nomenclature[A]{MDP}{Markov Decision Process}
\Nomenclature[A]{MZ VAV}{Multi Zone Variable Air Volume}
\Nomenclature[A]{OPC}{Open Platform Communication}
\Nomenclature[A]{PO}{Policy Optimisation}
\Nomenclature[A]{RBC}{Rule Based Controller}
\Nomenclature[A]{RL}{Reinforcement Learning}
\Nomenclature[A]{SCADA}{Supervisory Control and Data Acquisition}
\Nomenclature[A]{SAC}{Soft Actor Critic}
\Nomenclature[A]{TES}{Thermal Energy Storage}

\Nomenclature[G]{$\alpha$}{Temperature parameter of SAC (-)}
\Nomenclature[G]{$\beta$}{Power consumption weighting term (-)}
\Nomenclature[G]{$\gamma$}{Discount rate (-)}
\Nomenclature[G]{$\lambda$}{Comfort weighting term (-)}
\Nomenclature[G]{$\tau$}{SAC target smoothing coefficient (-)}
\Nomenclature[G]{$\pi$}{RL policy (-)}
\Nomenclature[G]{$\rho$}{Probability (-)}

\Nomenclature[R, 03]{$e(t)$}{Grid signal, e.g. real-time price (Euros/kWh)}
\Nomenclature[R, 10]{$S$}{A set of states (-)}
\Nomenclature[R, 01]{$A$}{A set of actions (-)}
\Nomenclature[R, 08]{$r$}{Reward function (-)}
\Nomenclature[R, 09]{$Q$}{Q-value (-)}
\Nomenclature[R, 06]{$P$}{Transition probabilities between states (-)}
\Nomenclature[R, 13]{$V^{\pi}(S)$}{Value function (-)}
\Nomenclature[R, 02]{$e_{HVAC}$}{HVAC power consumption (kWh)}
\Nomenclature[R, 12]{$t_{min}$}{Minimum temperature bound ($^{\circ} C$)}
\Nomenclature[R, 11]{$t_{max}$}{Maximum temperature bound ($^{\circ} C$)}
\Nomenclature[R, 05]{$J$}{Optimisation objective (-)}
\Nomenclature[R, 04]{$\mathcal{H}$}{Entropy (-)}

\Nomenclature[Su]{*}{Optimal}


\bibliography{J3_AK}

\end{document}